\pdfoutput=1

\documentclass[11pt]{article}

\usepackage[final]{acl}

\usepackage{times}
\usepackage{latexsym}

\usepackage[T1]{fontenc}

\usepackage[utf8]{inputenc}

\usepackage{microtype}

\usepackage{inconsolata}

\usepackage{graphicx}
\usepackage{amsmath}
\usepackage{xcolor}
\usepackage{booktabs}
\usepackage{geometry}
\usepackage{multicol}
\usepackage[most]{tcolorbox} 
\usepackage{caption}
\usepackage{placeins}
\usepackage{amssymb}%
\usepackage{pifont}%
\usepackage{floatrow}
\usepackage{fancyvrb}
\usepackage{graphicx}
\usepackage{amsmath}
\usepackage{amsfonts}       %
\usepackage{nicefrac}       %
\usepackage{xspace}         %
\usepackage{multirow}
\usepackage{array}
\usepackage{siunitx}
\usepackage{booktabs}
\usepackage{colortbl}
\usepackage{color}
\usepackage{soul}
\usepackage{highlight}
\usepackage{tcolorbox}
\usepackage{hyperref}
\usepackage{mdframed}
\PassOptionsToPackage{hyphens}{url}
\usepackage{enumitem}

\definecolor{casing}{RGB}{242,217,198}
\definecolor{entity}{RGB}{247,225,147}
\definecolor{negation}{HTML}{FFCCCC}
\definecolor{dialect}{HTML}{CCCCFF}
\definecolor{typos}{rgb}{.90,.92,1}
\definecolor{amplification}{rgb}{.91,.95,0.83}
\definecolor{pos}{RGB}{167, 199, 231}
\definecolor{neg}{RGB}{255, 120, 100}
\definecolor{rerank_pos}{RGB}{153, 220, 102}
\DeclareRobustCommand{\hlpos}[1]{{\sethlcolor{pos}\hl{#1}}}
\DeclareRobustCommand{\hlneg}[1]{{\sethlcolor{neg}\hl{#1}}}
\DeclareRobustCommand{\hlrerank}[1]{{\sethlcolor{rerank_pos}\hl{#1}}}
\DeclareRobustCommand{\hlbase}[1]{{\sethlcolor{lightgray}\hl{#1}}}
\DeclareRobustCommand{\hlpet}[1]{{\sethlcolor{negation}\hl{#1}}}
\tcbset{
  promptstyle/.style={
    enhanced,
    width=\textwidth,          %
    colback=gray!10,           %
    colframe=black,            %
    boxrule=0.5pt,             %
    arc=4pt,                   %
    left=6pt, right=6pt,       %
    top=6pt, bottom=6pt,       %
    fontupper=\ttfamily,       %
    colbacktitle=black,        %
    coltitle=white,            %
    title style={font=\bfseries, halign=flush left}, %
    boxed title style={arc=4pt},                   %
    attach boxed title to top left={xshift=0mm, yshift=-2mm}, %
  }
}
\newtcolorbox{prompt}[1][]{%
  promptstyle,
  title={Prompt},
  #1
}

\newcommand{\checkboxYes}{$\square$ Yes}
\newcommand{\checkboxNo}{$\square$ No}

\title{\textit{When Claims Evolve}: Evaluating and Enhancing the Robustness of Embedding Models Against Misinformation Edits}

\author{
 \textbf{Jabez Magomere,\textsuperscript{1}}
 \textbf{Emanuele La Malfa,\textsuperscript{1,2}}
 \textbf{Manuel Tonneau,\textsuperscript{1,3,4}} \\
 \textbf{Ashkan Kazemi,\textsuperscript{5}}
 \textbf{Scott A.\ Hale\textsuperscript{1,5}}
\\
 \textsuperscript{1}University of Oxford,
 \textsuperscript{2}Alan Turing Institute,
  \textsuperscript{3}World Bank, \\
 \textsuperscript{4}New York University, 
 \textsuperscript{5}Meedan
\\
\texttt{jabez.magomere@keble.ox.ac.uk}
}

\begin{document}
\maketitle
\begin{abstract}

Online misinformation remains a critical challenge, and fact-checkers increasingly rely on claim matching systems that use sentence embedding models to retrieve relevant fact-checks. However, as users interact with claims online, they often introduce edits, and it remains unclear whether current embedding models used in retrieval are robust to such edits. To investigate this, we introduce a perturbation framework that generates valid and natural claim variations, enabling us to assess the robustness of a wide-range of sentence embedding models in a multi-stage retrieval pipeline and evaluate the effectiveness of various mitigation approaches. Our evaluation reveals that standard embedding models exhibit notable performance drops on edited claims, while LLM-distilled embedding models offer improved robustness at a higher computational cost. Although a strong reranker helps to reduce the performance drop, it cannot fully compensate for first-stage retrieval gaps. To address these retrieval gaps,  we evaluate train- and inference-time mitigation approaches, demonstrating that they can improve in-domain robustness by up to 17 percentage points and boost out-of-domain generalization by 10 percentage points. Overall, our findings provide practical improvements to claim-matching systems, enabling more reliable fact-checking of evolving misinformation.
\end{abstract}

\section{Introduction}
The spread of misinformation, false or inaccurate information, is considered to be one of the biggest short-term threats to the cohesion within and between nations \cite{wef2025global}. %
While misinformation spreads fast \cite{vosoughi2018spread}, manual verification and debunking of false claims takes time. 

\begin{figure}[t]
    \centering
    \includegraphics[width=1.0\textwidth]{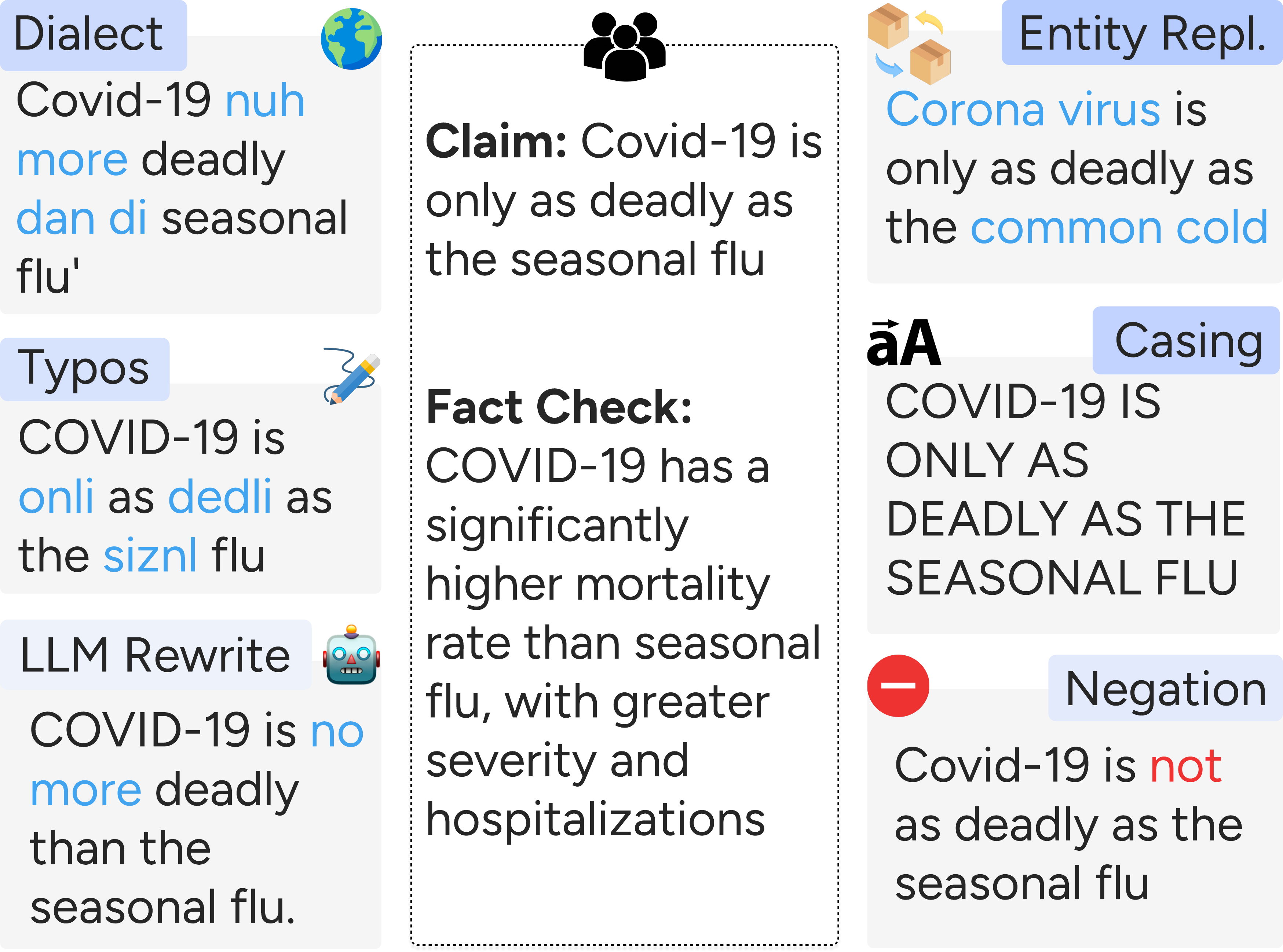}
    \caption{\textbf{Examples of user-informed misinformation edits observed in fact-checking}. Users frequently introduce subtle edits to online misinformation claims that preserve the corresponding fact-check. However, it remains unclear whether sentence embedding models are robust to these edits.}
    \label{fig:example_figure}
\end{figure}

\begin{figure*}[t]
    \centering
    \includegraphics[width=0.9\textwidth]{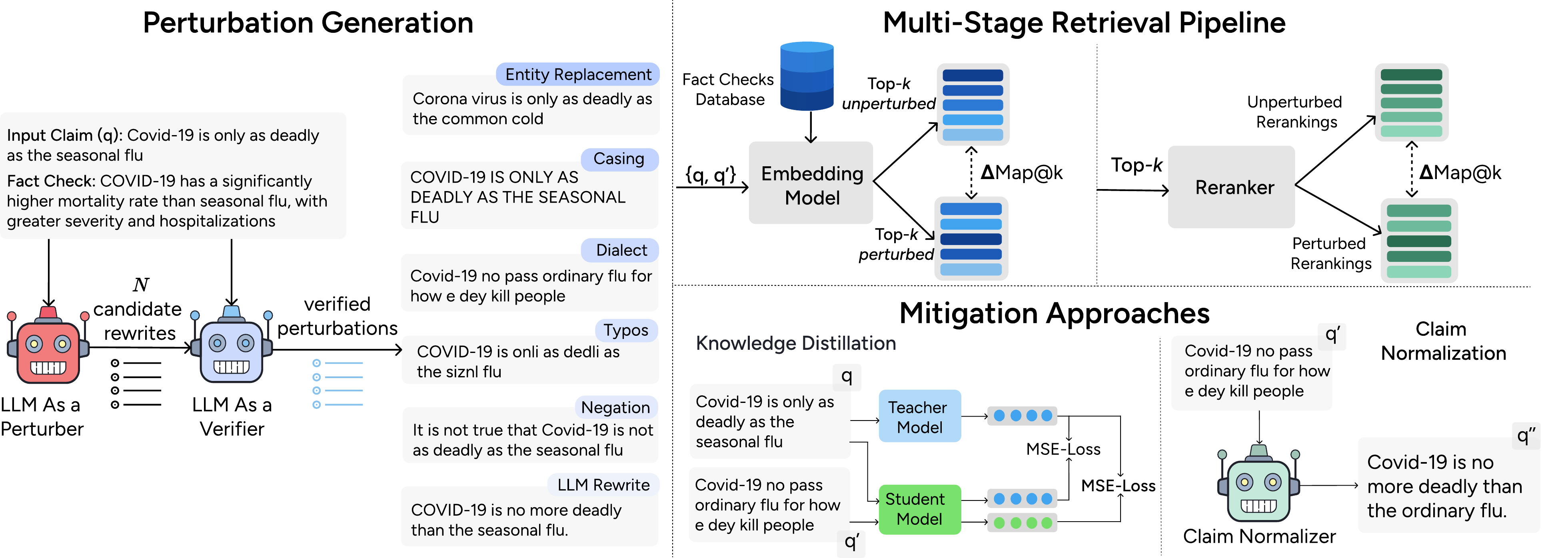}
    \caption{\textbf{An Overview of Our Approach}: (1) We generate candidate rewrites using an \textit{LLM As a Perturber}, followed by another \textit{LLM As a Verifier} of these perturbations. (2) Next, we assess the robustness of various embedding models within a multi-stage retrieval pipeline. (3) Finally, we examine the impact of train- and inference-time mitigation approaches on improving robustness for weaker embedding models.}
    \label{fig:overall_pipeline}
\end{figure*}

To speed up the fact-checking process, a growing body of research has focused on developing tools to help support human fact-checkers~\cite{nakov_automated_2021, guo-etal-2022-survey, das2023state}. One such approach is \textbf{claim matching}, the task of determining whether a given claim has already been fact-checked or retrieving relevant fact-checks ~\cite{shaar-etal-2020-known, sheng-etal-2021-article, shaar-etal-2022-assisting, kazemi-etal-2021-claim}. Claim matching has shown strong potential in reducing the time human fact-checkers spend verifying repeated or previously debunked claims, by ensuring that once a costly human fact-check is written, it can be linked to all relevant instances of that claim to provide context to others encountering it. However, current claim matching approaches are typically trained on datasets composed of previously fact-checked claims~\cite{barron2020overview, clef-checkthat:2022:task2}, excluding the vast majority of unverified claims in the wild. This limits models’ exposure to the noisy, diverse, and evolving nature of misinformation as it appears in real-world contexts. As a result, these models may perform well in controlled, in-domain evaluations but generalize poorly to settings characterized by unpredictable user behaviour.

As users interact with and share misinformation online, they often introduce various edits to claims, such as replacing entities, introducing common social-media slang, or rewriting a claim in a different dialect \cite{kazemi-etal-2023-query,quelle2025lost} as shown in Figure \ref{fig:example_figure}. While these edits may be subtle and not affect the matching fact-check, \citet{yan2022mutated} show that \textbf{mutated claims are more likely to spread widely on social media compared to non-mutated claims}. Thus, current claim-matching methods need to be robust to user claim edits to curb the spread of misinformation effectively. 

At the core of claim matching tasks is the reliance on sentence embedding models to represent input claims and fact-checks during retrieval. There has been growing interest in developing general-purpose sentence embeddings designed to handle a wide range of tasks \cite{lee2024geckoversatiletextembeddings,su-etal-2023-one}. However, their effectiveness in task-specific applications remains uncertain, as different domains demand varying notions of similarity \cite{ghafouri2024love}. Fact-checking is one such domain where fact-checkers need to retrieve sentences that may be evaluated under the same factual basis, even if they are not strictly semantically equivalent, a nuance that general-purpose embeddings mostly trained on semantic textual similarity datasets \cite{cer-etal-2017-semeval} may struggle to capture. 

To tackle the challenge of embedding model brittleness in real-world settings, we study \textbf{the robustness of sentence embedding models to user-driven misinformation edits} and \textbf{propose mitigation strategies to improve robustness}. We introduce a taxonomy of real-world edits drawn from fact-checking scenarios and evaluate model robustness on the downstream task of claim matching (Figure \ref{fig:overall_pipeline}). Our analysis shows that widely used embedding models are highly sensitive to user-informed misinformation edits, and that even strong rerankers fail to fully mitigate these weaknesses. However, we demonstrate that both train-time and inference-time mitigation approaches can significantly improve in-domain and out-of-domain robustness.  Based on these findings, our key contributions are as follows:\footnote{\href{https://github.com/JabezNzomo99/claim-matching-robustness}{Code and datasets} are available for research purposes.} 
\begin{itemize}[noitemsep] 
    \item We propose a novel perturbation framework for generating valid and natural claim edits, resulting in a new dataset for evaluating robustness of sentence embedding models. 
    \item We conduct a comprehensive evaluation of a wide range of embedding models, including BERT-based, T5-based, and LLM-distilled in a multi-stage retrieval pipeline comprising first-stage retrieval followed by reranking.
    \item We assess the effectiveness of both train-time and inference-time mitigation approaches in improving the robustness of weaker embedding models, providing insights into their trade-offs and effectiveness.
\end{itemize}

\section{Related Work}

\paragraph{Robustness in Misinformation Detection} Robustness in NLP is a well-studied area, with research focusing on developing adversarial attacks that can fool models and compromise real-world deployments \cite{malfa_king_2022, wang-etal-2022-measure, goyal2023survey}. In the context of misinformation, prior work has explored adversarial robustness in text classifiers for fake news detection \cite{zhou2019fake, le2020malcom, flores2022adversarial}, veracity prediction \cite{thorne-etal-2019-evaluating, hidey-etal-2020-deseption}, and propaganda detection \cite{Przybyła_Shvets_Saggion_2024}. However, fact checkers have evidenced a lack of transparency and credibility on automated fact-checking approaches that assign factuality labels to claims \cite{glockner-etal-2022-missing, das2023state, procter2023some}. In contrast, the robustness of \textit{claim matching}---a core information retrieval task---remains underexplored, with prior research focusing on improving retrieval performance \cite{barron2020overview, sheng-etal-2021-article, nakov2021clef, clef-checkthat:2022:task2}. Most studies on robustness in misinformation rely on adversarial techniques, such as character- and word-level substitutions \cite[e.g., synonym replacements: ][]{jin2019bert, li-etal-2020-bert-attack, garg-ramakrishnan-2020-bae}, to perturb input claims. However, the practicality of these techniques has been questioned, as recent studies show that such attacks can mislead NLP models but have little impact on human readers \cite{morris-etal-2020-reevaluating, dyrmishi-etal-2023-humans, chiang-lee-2023-synonym}. There is also a well-known trade-off between how realistic a local adversarial attack is and its plausibility for humans \cite{malfa_king_2022}. This issue is even more critical in misinformation contexts, where the goal is to not only fool a model but also to generate believable claims and spread content among human users; therefore, the perturbations should be both valid and natural \cite{dyrmishi-etal-2023-humans}. In contrast to adversarial robustness which focuses on exploiting model vulnerabilities, our work focuses on \textit{realistic user edits}: natural variations in misinformation that arise organically, either intentionally or unintentionally, as claims spread online. 

\paragraph{Evaluating Embedding Model Robustness} Several studies have examined the robustness of embedding models across different settings. \citet{rafiei-asl-etal-2024-robustsentembed} evaluate robustness of embedding models adversarially using TextAttack~\cite{morris2020textattack} and propose using gradient-based~\cite{goodfellow2014explaining} perturbed embeddings to enhance robustness. \citet{chiang-etal-2023-revealing} examine how different sentence encoders represent sentence pairs with high lexical overlap, showing that training datasets influence similarity notions. Closely related to our work, \citet{nishimwe-etal-2024-making} assess \texttt{LASER} \cite{heffernan-etal-2022-bitext} against user-generated content and apply knowledge distillation ~\cite{reimers-gurevych-2020-making} to improve robustness. However, prior work largely evaluates embeddings in isolation, despite their role in larger systems: e.g., candidate selection in multi-stage retrieval~\cite{nogueira-etal-2020-document} or retrieval-augmented generation (RAG) pipelines~\cite{fan2024survey}. The impact of representation differences on end-to-end system performance remains unclear. For example, we examine whether the robustness of embedding models in first-stage retrieval meaningfully affects the overall performance of a multi-stage retrieval system when a strong reranker is applied, addressing this gap.

\section{Data and Methods}
\subsection{Background}
\subsubsection{Task Definition}
Following \citet{shaar-etal-2020-known}, we define the task of claim matching as a \emph{learning-to-rank problem}. Given a check-worthy input claim \( q \) and a set of verified claims or fact-checks \( V = \{ v_1, v_2, \dots, v_n \} \), the objective is to learn a ranking function \( f \) that orders the verified claims such that those that can help verify the input claim \( q \) are ranked higher. \( f \) returns a ranked list \( [ v_{\pi(1)}, v_{\pi(2)}, \dots, v_{\pi(n)} ] \), where \( \pi \) is a permutation of indices that sorts the claims in descending order of relevance to \( q \). 
Model performance on this task is evaluated using Mean Average Precision (MAP) score truncated to rank \(k\) (MAP@\(k\)). We compare claims with a sentence embedding model \( \theta \) that encodes the input claim and fact-checks into dense vector representations. These embeddings are then utilized in learning the ranking function \( f \).

\subsubsection{Problem Formulation}
\label{sec:problem_formulation}
We consider a scenario where a social media user $u$ edits an original claim $q$ to produce a perturbed claim $q'$. The edits must ensure that $q'$ is both \textit{\textbf{valid}}, remaining applicable to the fact-check $v$ and conveying the same claim as $q$ and \textit{\textbf{natural}}, meaning that $q'$ is something a social media user might write. These edits are modeled as a set of transformations $\mathcal{T}$, each applied with a noise level $b$ drawn from a budget $B_t = [b_{\text{low}}^t,\, b_{\text{high}}^t]$ for $t \in \mathcal{T}$. We vary the noise levels to capture the spectrum of user edits, from minimal edits (low noise) to substantial rewording (high noise). A set of constraints $\mathcal{C}$ enforces the validity and naturalness of $q'$. Formally, this process is described by
\begin{equation}
q' = t_b(q) \quad \text{with} \quad \mathcal{C}(q,q',v) = \text{True}
\end{equation}
where given an input claim \( q \), the \emph{baseline edit} is defined as $q_{\text{base}}^t = t_{b_{\text{low}}^t}(q)$ and the \emph{worst-case edit} as $q_{\text{worst}}^t = t_{b_{\text{high}}^t}(q)$. This formulation systematically examines how varying noise levels within $B_t$ impact sentence embedding models in claim matching tasks.

\subsubsection{Perturbations as Claim Edits}
We introduce a taxonomy of common misinformation edits observed in real-world fact-checking, which serves as a template for the transformations \( t \in \mathcal{T} \). The applied transformations and noise levels are as follows:

\paragraph{Casing} A subtle transformation observed in claims involves changes in casing, such as uppercasing an entire claim or parts of it to add emphasis. 
In other cases, casing differences may be artefacts of the noisy nature of social media text \cite{ritter2011named, baldwin2015shared}.
To capture casing variations, we apply \texttt{TrueCasing} \cite{lita2003truecasing}\footnote{A preprocessing technique that converts all-lowercase text to properly capitalized text and introduces variations in text.} to the original claim as the minimum amount of noise (\( b_{\text{low}} \)) and \texttt{UPPERCASE} the entire input claim as the maximum noise level (\( b_{\text{high}}\)).

\paragraph{Typos}
User edits to a claim may introduce spelling errors, common social media abbreviations and slang \cite{van2018taxonomy, sanguinetti2020treebanking}. We define a transformation budget based on the \textit{Levenshtein} edit distance between the unperturbed and perturbed input claims. The baseline budget, \(b_{\text{low}}\), corresponds to the perturbed claim with the smallest edit distance, while the worst-case budget, \(b_{\text{high}}\), represents the largest edit distance to the unperturbed claim.

\paragraph{Negation} 
Social media users may negate claims, for instance as part of counterspeech~\cite{gligoric-etal-2024-nlp}, though the underlying claim to be verified remains unchanged. To evaluate embedding models on such cases, we apply negation transformations to original claims, defining the minimum noise level (\(b_{\text{low}}\)) as a single negation and the maximum noise level (\(b_{\text{high}}\)) as a double negation. For example, the claim ``Covid-19 is only as deadly as the seasonal flu'' with a single negation becomes ``Covid-19 is not as deadly as the seasonal flu''. With double negation the same claim becomes ``It is not true that Covid-19 is not as deadly as the seasonal flu''.

\paragraph{Entity Replacement} 
One common pattern in mutated claims is the substitution of entities in the original text with synonymous references, such as replacing a well-known person’s name with a nickname or changing entities to tailor a claim to a new audience; for example, changing “According to a Harvard study \ldots” to “According to an Oxford study \ldots” while retaining the original meaning. 
We define the transformation budget as the number of named entities to be replaced in a claim, with \( b_{\text{low}} = 1 \), meaning at most one entity is replaced, and \( b_{\text{high}} = All \), meaning all entities are replaced.

\paragraph{LLM rewrites} 
\citet{chen2024llmgenerated} demonstrate that LLMs can be used to rewrite a claim, either to conceal authorship or to make the claim more deceptive and harder to detect. We prompt an LLM to rewrite an original claim, setting the transformation budget based on the \textit{Levenshtein} distance, where \( b_{\text{low}} \) is the rewrite with the fewest changes, and \( b_{\text{high}} \) is the rewrite with the most changes.

\paragraph{Dialect Changes} 
A social media user may rewrite a claim in a different dialect to fit a specific target audience. However, most NLP models are primarily trained on American and British English \cite{longpre2024bridging,tonneau-etal-2024-languages}, leading to notable performance drops when applied to other English dialects \cite{jurgens-etal-2017-incorporating,ziems2023multi, tonneau-etal-2024-naijahate}. To study the effect of dialectal variations, we rewrite claims in \textit{African American English}, \textit{Nigerian Pidgin English}, \textit{Singlish} (an English variant spoken in Singapore) and \textit{Jamaican Patois}, an English-based creole spoken in the Caribbean.\footnote{We treat all dialects equal and assume similar noise levels.}

\subsection{Experimental Setup}
\subsubsection{Perturbation Generation Framework}
\label{sec:generationframework}
To generate perturbations that meet the constraints outlined in Section \S\ref{sec:problem_formulation}, we implement a two-stage, multi-prompting framework leveraging two separate LLMs: one for perturbing input claims and another for verifying them. In the \textbf{perturbation step}, the \textit{perturber} LLM takes an input claim $q$ and its corresponding fact-check $v$ as inputs to generate an initial set of $N$ candidate rewrites intended to satisfy the transformation requirements and constraints. To select valid perturbations, we implement a \textbf{verification step}, where another LLM assumes the role of \textit{verifier} and checks whether the generated rewrites adhere to all constraints. Each perturbation is assigned a binary label ($1$ for valid, $0$ for invalid), and invalid perturbations are discarded. We set the initial candidate size $N$ to $5$ and use \texttt{GPT4o} as both the \textit{perturber} and \textit{verifier}. A human evaluation of the perturbation framework shows that the \textit{perturber} applies perturbations with 96.70\% accuracy, while the \textit{verifier} achieves 100\% precision in selecting valid perturbations. However, the verifier’s lower recall of 55.67\% suggests that it may be overly conservative, potentially filtering out some valid instances. 
Further details on the human evaluation and prompts are provided in Sections~\S\ref{appendix:human_evaluation} and \S\ref{appendix:prompt}, respectively.

\subsubsection{Datasets}
We apply perturbations to two claim-matching datasets: \textit{CheckThat22} \cite{clef-checkthat:2022:task2} and the dataset proposed by \citet{kazemi2022matching}, referred thereafter as \textit{FactCheckTweet}. \textit{CheckThat22} matches tweets with short, verified claims, whereas \textit{FactCheckTweet} matches claims with full fact-check articles. We apply perturbations to the test splits only (see Appendix \S\ref{appendix:dataset_statistics}). To assess generalization i.e. ability to improve performance on novel, out-of-distribution claims, we test our mitigation approaches on an out-of-domain (\textit{OOD Dataset}) composed of novel claim--fact-check pairs provided by Meedan\footnote{The non-profit Meedan supports fact-checkers running misinformation tiplines on WhatsApp and other platforms.} and primarily sourced through five fact-checking organizations. Table~\ref{tab:dataset_stats} summarizes the descriptive statistics of these datasets. \begin{table}[ht]
\centering
\resizebox{1.0\textwidth}{!}{%
\begin{tabular}{l|rr|rr}
\toprule
\textbf{Dataset} & \textbf{\# Claim} & \textbf{\# FC} & \textbf{Claim Len} & \textbf{FC Len} \\
\midrule
\textit{CheckThat22}   & 2,592 & 14,231 & 42.3  & 19.2  \\
\textit{FactCheckTweet} & 3,578 & 59,951 & 37.9  & 682.1 \\
\textit{OOD Dataset}    & 721   & 1,261  & 150.1 & 80.4  \\
\bottomrule
\end{tabular}%
}
\caption{Statistics of datasets in this work. FC denotes target fact-checks; lengths are measured in words.}
\label{tab:dataset_stats}
\end{table}
\vspace{-1ex}

\subsubsection{Multi-Stage Retrieval Pipeline}
To evaluate the robustness of various embedding models, we adopt a common multi-stage retrieval pipeline \cite{nogueira1901passage, nogueira2019multi, ma2024fine} consisting of a computationally efficient \textit{retriever} model to return top\textit{-j} relevant fact-checks followed by a more computationally intensive \textit{reranker} that refines the retrieved candidates to improve ranking quality. We describe the models we evaluate for each stage below:

\paragraph{Retrievers} 
In the retrieval stage, we evaluate embedding models with varying dimensions, pre-training approaches, and backbone architectures. This includes: \textbf{BERT-based embedding models}~\cite{devlin-etal-2019-bert, reimers-gurevych-2019-sentence}: 
\texttt{all-MiniLM-L12-v2}, \texttt{all-mpnet-base-v2}, and \texttt{all-distilroberta-v1}. \textbf{T5-based embedding models}~\cite{ni-etal-2022-sentence, raffel2020exploring}: \texttt{sentence-t5-base}, \texttt{sentence-t5-large}, \texttt{instructor-base}, and \texttt{instructor-large}—instruction-finetuned embeddings based on GTR models~\cite{su-etal-2023-one}. \textbf{Decoder-only LLM-distilled embeddings}: \texttt{SFR-Embedding-Mistral}~\cite{SFRAIResearch2024} and \texttt{NV-Embed-v2}~\cite{lee2025nvembedimprovedtechniquestraining},\footnote{\texttt{NV-Embed-v2} was the top-performing model on MTEB~\cite{muennighoff2022mteb} at the time of this study. \url{https://huggingface.co/spaces/mteb/leaderboard}} both initialized from \texttt{Mistral-7B}. To study the effect of finetuning, we finetune \texttt{all-mpnet-base-v2} and \texttt{sentence-t5-large} on the \textit{CheckThat22} training set, and denote the finetuned models respectively as \texttt{all-mpnet-base-v2-ft} and \texttt{sentence-t5-large-ft}. We also implement \textbf{BM25 as a baseline} to explore the robustness of lexical-based sparse retrieval approaches. We provide further details on embedding models and finetuning in \S\ref{appendix:embedding_models} of the Appendix. 

\paragraph{Rerankers}  
With multiple combinations of embedding models (\( m = 12 \)), perturbation types (\( p = 14 \)), and rerankers (\( r = 7 \))---totalling \( m \times p \times r = 1,176\) cases, we evaluate a wide range of rerankers and select the best-performing one for further experiments.  While \texttt{RankGPT}~\cite{Sun2023IsCG} using \texttt{GPT4o} achieves the highest performance, we select \texttt{bge-reranker-v2-gemma}~\cite{chen2024bge} for its comparable accuracy and significantly lower inference cost. See \S\ref{appendix:selecting_strong_reranker} in the Appendix for reranker evaluations.

\section{Results}
\subsection{RQ1: How do perturbations affect the robustness of embedding models in first-stage retrieval?}
\label{sec:retrieval_results}
\begin{table*}[t!]
\centering
\resizebox{0.9\textwidth}{!}{%
\begin{tabular}{ll|cc|cc|cc|cc|cc|cccc}
\toprule
 \multicolumn{2}{l}{} & \multicolumn{2}{c|}{\textbf{Casing}} & \multicolumn{2}{c|}{\textbf{Typos}} & \multicolumn{2}{c|}{\textbf{Negation}} & \multicolumn{2}{c|}{\textbf{Entity Replacement}} & \multicolumn{2}{c|}{\textbf{LLM Rewrite}} & \multicolumn{4}{c}{\textbf{Dialect}} \\
 \textbf{Stage} & \textbf{Model} & \textit{TrueCase} & \textit{Upper} & \textit{Least} & \textit{Most} & \textit{Shallow} & \textit{Double} & \textit{Atleast 1} & \textit{All} & \textit{Least} & \textit{Most} & \textit{AAE} & \textit{Jamaican} & \textit{Pidgin} & \textit{Singlish} \\
\midrule
\multicolumn{15}{c}{\large\bfseries \textit{CheckThat22}} \\
\bottomrule
\end{tabular}}
\caption{Effect of perturbations on \textbf{\textit{CheckThat22}} (top) and \textbf{\textit{FactCheckTweet}} (bottom). \textbf{First-Stage Retrieval} shows the \(\Delta \text{\emph{retrieval gap}}\) between unperturbed and perturbed inputs sets. \textbf{Reranking Recovery} measures the \(\Delta \text{\emph{recovery gap}}\) for the perturbed input set before and after reranking top-$50$ candidates, with improvements highlighted in \hlrerank{green} and drops in \hlneg{red}. \textbf{Overall Pipeline} shows the \(\Delta \text{\emph{overall gap}}\) for combined retrieval and reranking between unperturbed and perturbed inputs. Colors indicate \hlpos{positive} or \hlneg{negative} deltas compared to unperturbed. \(\text{MAP\textit{@}}20\) is used for all comparisons. Models finetuned are indicated with the postfix \texttt{ft}. \vspace{-0.5em}}
\label{tab:map@20_checkthat}
\end{table*}

We begin by evaluating the robustness of embedding models used for first-stage retrieval under various perturbation types. For each embedding model and perturbation type, we compute the \(\Delta \text{\emph{retrieval gap}}\) by comparing the MAP@\textit{k} performance difference between the unperturbed and perturbed sets for a given value of \textit{k}. Results for \textit{CheckThat22} and \textit{FactCheckTweet}  with \( k = 20 \) are shown in Table~\ref{tab:map@20_checkthat}. We also provide results for additional values of \( k \), which exhibit similar trends, in Appendix \S\ref{appendix:additional_results}. From these results, we make the following observations:

\paragraph{LLM-distilled embeddings exhibit superior robustness.} Our analysis shows a smaller \(\Delta \text{\emph{retrieval gap}}\) for \texttt{SFR-Embedding-Mistral} and \texttt{NV-Embed-V2}, both based on \texttt{Mistral-7B} \cite{jiang2023mistral7b}. We hypothesize that this robustness stems from the increased representational capacity afforded by the larger scale of the backbone model from which the embeddings are derived, a characteristic shown to enhance robustness to adversarial inputs \cite{howe2024effectsscalelanguagemodel}. This robustness also extends to dialectical variations and social-media typos, which may be due to the increasing importance of social media data in the training corpora of recent generative models \cite{longpre2024bridging, miranda2024scalediversitycoefficientdata, penedo2024the}. In contrast, BERT-based and T5-based models are primarily trained on less diverse, standardised corpora, making them less robust to lexical variations. 

\paragraph{NLI-Fine-Tuned embedding models such as Sentence-T5 demonstrate low robustness to negation.} Sentence embedding models primarily trained on natural language inference (NLI) datasets, such as T5 \cite{raffel2020exploring}, exhibit a large \(\Delta \text{\emph{retrieval gap}}\) under negation. This may stem from treating contradiction hypotheses as hard-negatives during training with contrastive loss \cite{ni-etal-2022-sentence}, making the models highly sensitive to negation \cite{chiang-etal-2023-revealing}. In contrast, models trained on more diverse datasets (e.g., \texttt{all-distilroberta-v1}, \texttt{all-MiniLM-L12-v2}, \texttt{all-mpnet-base-v2}) demonstrate greater robustness to \textit{negation} and, in some cases, even show improvements. BM25, meanwhile, remains robust to negation, as it is affected minimally by small lexical changes. Finally, \textit{double negation} rewrites claims in a form closely aligned with fact-checks (e.g., ``It is not true \ldots''), and may lead to retrieval improvements for some embedding models as observed on \textit{FactCheckTweet}.

\paragraph{Embedding models with case-sensitive tokenizers struggle with casing variations.} Embedding models such as \texttt{all-distilroberta-v1} and \texttt{T5}-based models with case-sensitive tokenizers, which distinguish between tokens like \texttt{apple} and \texttt{Apple}, show significant retrieval performance gaps on casing-perturbed claims. In contrast, embedding models with case-insensitive tokenizers exhibit robustness to casing differences, showing minimal to no retrieval gaps. While this behavior is expected for case-sensitive models, in retrieval tasks where casing differences are irrelevant, applying casing normalization or using a case-insensitive model can be an effective solution.

\subsection{RQ2: Can a Strong Reranker Recover the Retrieval Gap?}
\label{sec:reranking_results}
In Section \S\ref{sec:retrieval_results}, we observe that perturbations introduce significant retrieval gaps in first-stage retrieval. However, it remains unclear whether a strong reranker can mitigate these gaps and improve ranking quality. To test this, we apply a reranker to the top-\(j\) retrieved candidates and measure the \(\Delta \text{\emph{recovery gap}}\), which quantifies the MAP@\textit{k} improvement on the perturbed set after reranking. We rerank the top \(j=50\) candidates, as this balances efficiency and performance, achieving the highest MAP@$5$ across evaluated values (\(j \in \{5, 10, 20, 50, 100, 200\}\)) (see \S\ref{appendix:selecting_optimal_k}). We report this gap for \(k=20\) and \(j=50\) on \textit{CheckThat22} and \textit{FactCheckTweet} using the reranker \texttt{bge-reranker-v2-gemma} in Table~\ref{tab:map@20_checkthat}. Additional results for different \(k\) values are provided in \S\ref{appendix:additional_results} of the Appendix. Across different values of \(k\) and evaluation datasets, we observe that:

\paragraph{Reranking improves weaker embedding models but can negatively impact stronger models in short-target tasks.}
Across both datasets, we observe that reranking is largely beneficial for embedding models that are less robust in first-stage retrieval. However, in \textit{CheckThat22}, which contains shorter targets (i.e., short fact-check statements), reranking reduces performance for stronger embedding models, as shown by a drop in MAP@$20$ scores. This decline may stem from noise i.e., inclusion of additional, potentially irrelevant or low-quality candidates introduced when reranking a larger set of top-$50$ candidates, which can distort the high-quality rankings produced by stronger models.
For stronger embedding models, the most relevant candidates are often already ranked highly and expanding the candidate set can introduce noise. When using stronger embedding models in first-stage retrieval, reranking a smaller candidate subset or skipping reranking altogether might mitigate this issue. For \textit{FactCheckTweet}, the reranker processes the full fact-check article due to its longer context window, unlike the paragraph-based splitting used during retrieval. This allows the reranker to leverage the full fact-check's information, which we hypothesize contributes to its more pronounced reranking gains, even for stronger models, compared to \textit{CheckThat22}.

\subsection{RQ3: To what extent do perturbations impact the overall pipeline's performance?}
We observe that a strong reranker is helpful in recovering the retrieval gap discussed in Section \S\ref{sec:reranking_results}. However, whether it fully compensates for this gap remains to be determined. Here, we examine the end-to-end pipeline (i.e., retrieval plus reranking), by comparing performance on perturbed vs.\ unperturbed sets. This analysis reveals the \(\Delta \text{\emph{overall gap}}\), which is the MAP@$k$ difference between the unperturbed and perturbed sets for the full pipeline of retrieval and reranking the \textit{top-}$j$ candidates. We show the results in Table \ref{tab:map@20_checkthat} for $k=20, j=50$, and provide additional results for other $k$ values in \S\ref{appendix:additional_results} of the Appendix. We observe that:

\paragraph{A strong reranker helps, but cannot fully compensate for first-stage retrieval gaps.} A strong reranker proves effective in reducing the retrieval gap particularly for less robust embedding models. However, a notable end-to-end performance gap persists, indicating that a strong reranker alone is not sufficient. Although we select a SoTA reranker, we hypothesize that it may also exhibit brittleness to some of the perturbations, which could reduce its effectiveness in reranking and contribute to the observed performance drop. For instance, under the \textit{negation} perturbation on \textit{CheckThat22}, the reranker negatively impacts the overall performance of embedding models that initially performed well in first-stage rankings. Additionally, weaker embedding models display significant deficiencies in handling \textit{typos} and \textit{dialect} perturbations. Similar limitations are observed with case-sensitive embedding models, where reranking fails to fully recover from retrieval gaps caused by \textit{casing} perturbations. For \textit{entity replacements}, reranking helps reduce the retrieval gap on \textit{CheckThat22}, but offers little benefit on \textit{FactCheckTweet}. We hypothesize that entity swaps have a smaller impact when matching against shorter targets, as in \textit{CheckThat22}, compared to longer targets, where the effects of entity changes are more severe. 

\paragraph{\textit{LLM Rewrites} improves overall performance} as it acts a form of query rewriting or naïve query expansion, a technique shown to improve performance in retrieval tasks \cite{nogueira2019document, kazemi-etal-2023-query, weller-etal-2024-generative}.

\begin{table}[t!]
\centering
\resizebox{1.0\textwidth}{!}{%
\begin{tabular}{l|cc|cc|cc|cc}
\toprule
 \multicolumn{1}{l}{} & \multicolumn{2}{c|}{\textbf{\textit{Typos}}} & \multicolumn{2}{c|}{\textbf{\textit{Neg.}}} & \multicolumn{2}{c|}{\textbf{\textit{Entity R.}}} & \multicolumn{2}{c}{\textbf{\textit{Dialect}}} \\
 \textbf{Model} & \texttt{U} & \texttt{P} & \texttt{U} & \texttt{P} & \texttt{U} & \texttt{P} & \texttt{U} & \texttt{P} \\
\midrule
\texttt{mpnet (baseline)}         & \gradientcell{0.76}{0.74}{0.98}{low}{high}{\opacity}{0} & 0.68 & \gradientcell{0.78}{0.74}{0.98}{low}{high}{\opacity}{0} & 0.74 & \gradientcell{0.76}{0.74}{0.98}{low}{high}{\opacity}{0} & 0.66 & \gradientcell{0.74}{0.74}{0.98}{low}{high}{\opacity}{0} & 0.65 \\
\texttt{mpnet-robust}              & \gradientcell{0.83}{0.74}{0.98}{low}{high}{\opacity}{0} & 0.73 & \gradientcell{0.84}{0.74}{0.98}{low}{high}{\opacity}{0} & 0.81 & \gradientcell{0.83}{0.74}{0.98}{low}{high}{\opacity}{0} & 0.77 & \gradientcell{0.81}{0.74}{0.98}{low}{high}{\opacity}{0} & 0.80 \\
\texttt{mpnet-ft}                  & \gradientcell{0.90}{0.74}{0.98}{low}{high}{\opacity}{0} & 0.77 & \gradientcell{0.90}{0.74}{0.98}{low}{high}{\opacity}{0} & 0.85 & \gradientcell{0.89}{0.74}{0.98}{low}{high}{\opacity}{0} & 0.80 & \gradientcell{0.88}{0.74}{0.98}{low}{high}{\opacity}{0} & 0.80 \\
\texttt{mpnet-robust-ft}           & \gradientcell{0.89}{0.74}{0.98}{low}{high}{\opacity}{0} & 0.78 & \gradientcell{0.88}{0.74}{0.98}{low}{high}{\opacity}{0} & 0.84 & \gradientcell{0.90}{0.74}{0.98}{low}{high}{\opacity}{0} & 0.81 & \gradientcell{0.87}{0.74}{0.98}{low}{high}{\opacity}{0} & 0.84 \\
\midrule
\texttt{mpnet+CN}                & \gradientcell{0.77}{0.74}{0.98}{low}{high}{\opacity}{0} & 0.78 & \gradientcell{0.80}{0.74}{0.98}{low}{high}{\opacity}{0} & 0.79 & \gradientcell{0.76}{0.74}{0.98}{low}{high}{\opacity}{0} & 0.75 & \gradientcell{0.74}{0.74}{0.98}{low}{high}{\opacity}{0} & 0.73 \\
\texttt{mpnet-robust+CN}         & \gradientcell{0.83}{0.74}{0.98}{low}{high}{\opacity}{0} & 0.83 & \gradientcell{0.89}{0.74}{0.98}{low}{high}{\opacity}{0} & 0.82 & \gradientcell{0.84}{0.74}{0.98}{low}{high}{\opacity}{0} & 0.84 & \gradientcell{0.82}{0.74}{0.98}{low}{high}{\opacity}{0} & 0.82 \\
\texttt{mpnet-ft+CN}             & \gradientcell{0.89}{0.74}{0.98}{low}{high}{\opacity}{0} & \textbf{0.89} & \gradientcell{0.85}{0.74}{0.98}{low}{high}{\opacity}{0} & \textbf{0.87} & \gradientcell{0.89}{0.74}{0.98}{low}{high}{\opacity}{0} & 0.87 & \gradientcell{0.88}{0.74}{0.98}{low}{high}{\opacity}{0} & 0.86 \\
\texttt{mpnet-robust-ft+CN}      & \gradientcell{0.90}{0.74}{0.98}{low}{high}{\opacity}{0} & 0.88 & \gradientcell{0.90}{0.74}{0.98}{low}{high}{\opacity}{0} & 0.85 & \gradientcell{0.89}{0.74}{0.98}{low}{high}{\opacity}{0} & \textbf{0.89} & \gradientcell{0.89}{0.74}{0.98}{low}{high}{\opacity}{0} & \textbf{0.87} \\
\midrule
\texttt{NV-Embed-v2}             & \gradientcell{0.98}{0.74}{0.98}{low}{high}{\opacity}{0} & 0.98 & \gradientcell{0.97}{0.74}{0.98}{low}{high}{\opacity}{0} & 0.94 & \gradientcell{0.97}{0.74}{0.98}{low}{high}{\opacity}{0} & 0.97 & \gradientcell{0.97}{0.74}{0.98}{low}{high}{\opacity}{0} & 0.96 \\
\texttt{NV-Embed-v2 + CN}          & \gradientcell{0.98}{0.74}{0.98}{low}{high}{\opacity}{0} & 0.97 & \gradientcell{0.97}{0.74}{0.98}{low}{high}{\opacity}{0} & 0.93 & \gradientcell{0.97}{0.74}{0.98}{low}{high}{\opacity}{0} & 0.97 & \gradientcell{0.97}{0.74}{0.98}{low}{high}{\opacity}{0} & 0.97 \\
\bottomrule
\end{tabular}%
}
\caption{\textbf{Effect of mitigation approaches} on \textit{typos}, \textit{negation}, \textit{entity replacement} and dialect (\textit{Pidgin}) on \textit{\textbf{CheckThat22}} \textbf{(In-Domain)}. \texttt{mpnet} refers to \texttt{all-mpnet-base-v2}, with suffixes denoting mitigation approaches: \texttt{robust} (KD approach), \texttt{ft} (task finetuning), and \texttt{+CN} (claim normalization). Columns marked \texttt{U} and \texttt{P} indicate MAP@$20$ performance on the unperturbed and perturbed sets, respectively.}
\label{tab:refactored_mitigation_results}
\end{table}

\subsection{RQ4: How effective are mitigation approaches at improving robustness for weaker embedding models, and what trade-offs do they entail?}
LLM-distilled embeddings are more robust to misinformation edits (Section \S\ref{sec:retrieval_results}), but they come with the trade-off of increased storage requirements and slower inference, making them less ideal for first-stage retrieval. In contrast, smaller models are computationally efficient but exhibit weaker robustness. In this section, we explore whether train-time and inference-time approaches can improve the robustness of weaker embedding models on \textit{typos}, \textit{negation}, \textit{dialect changes}, and \textit{entity replacement}, perturbations that remain challenging even for a strong reranker. We test the approaches on the \textit{CheckThat22} dataset, which we refer to as the in-domain dataset, as we use its training and development sets. We also evaluate how well these improvements generalize to an out-of-domain (OOD) dataset consisting of novel claims not seen during training on \textit{CheckThat22}. All mitigations are applied only to the first-stage retrieval, and we report retrieval performance for this stage.\footnote{We assume that improving the first-stage retrieval will subsequently improve reranking performance.}

For train-time interventions, we examine \textbf{fine-tuning} on the \textit{CheckThat22} training set and a \textbf{knowledge distillation (KD)} approach inspired by \citet{reimers-gurevych-2020-making}, which aims to reduce representation differences between unperturbed and perturbed input claims. For KD, we generate parallel sentence pairs using our perturbation generation framework (\S\ref{sec:generationframework}), applied to the \textit{CheckThat22} training set, yielding 11,593 unperturbed–perturbed claim pairs. We further expand this set by pairing different perturbations of the same claim (e.g., typo and dialect), resulting in 70,954 claim pairs. Details of the training setup are provided in \S\ref{appendix:teacher_student}. As an inference-time intervention, we investigate a \textbf{claim normalization} approach proposed by \citet{sundriyal-etal-2023-chaos}, where input claims are rewritten into a standard form using \texttt{GPT4o} prior to retrieval.\footnote{The prompt used is shown in \S\ref{appendix:prompt}.} Finally, we evaluate the effect of combining multiple approaches. We select \texttt{all-mpnet-base-v2} as a representative weaker---yet computationally efficient---embedding model, as it is widely used for retrieval tasks.\footnote{\href{https://huggingface.co/sentence-transformers/all-mpnet-base-v2}{\texttt{all-mpnet-base-v2}} had 33.4M downloads in the past month at the time of this study.} Results are shown in Table \ref{tab:refactored_mitigation_results} and Table \ref{tab:ood_results} for the test split of \textit{CheckThat22} and the \textit{OOD Dataset}, respectively. We observe that:

\begin{table}[t!]
\centering
\resizebox{0.9\textwidth}{!}{%
\begin{tabular}{l|cccc}
\toprule
\textbf{Model} & MAP@$1$ & MAP@$5$ & MAP@$20$ & MAP@$50$\\
\midrule
\texttt{mpnet (baseline)}          & 0.4327 & 0.5129 & 0.5254 & 0.5269 \\
\texttt{mpnet-robust}               & 0.4508 & 0.5259 & 0.5390 & 0.5411 \\
\texttt{mpnet-ft}                   & 0.5076 & 0.5853 & 0.5982 & 0.5998 \\
\texttt{mpnet-robust-ft}            & 0.5090 & 0.5851 & 0.5991 & 0.6006 \\ \midrule
\texttt{mpnet+CN}                   & 0.4938 & 0.5696 & 0.5799 & 0.5816 \\
\texttt{mpnet-robust+CN}            & 0.5076 & 0.5765 & 0.5888 & 0.5911 \\
\texttt{mpnet-ft+CN}                & \textbf{0.5368}$^\dagger$ & \textbf{0.6109}$^\dagger$ & \textbf{0.6213}$^\dagger$ & \textbf{0.6233}$^\dagger$ \\
\texttt{mpnet-robust-ft+CN}         & 0.5257 & 0.6046 & 0.6163 & 0.6185 \\ \midrule
\texttt{NV-Embed-v2}                & 0.5520 & 0.6306 & 0.6402 & 0.6418 \\ 
\texttt{NV-Embed-v2+CN}             & \textbf{0.5576}$^*$ & \textbf{0.6375}$^*$ & \textbf{0.6464}$^*$ & \textbf{0.6479}$^*$ \\
\bottomrule
\end{tabular}%
}
\caption{\textbf{Effect of Mitigation Approaches on an Out-of-Domain (OOD) Dataset}. \texttt{mpnet} denotes \texttt{all-mpnet-base-v2}, with \texttt{-ft} for task finetuning, \texttt{-robust} for the KD approach, and \texttt{+CN} indicating claim normalization. $^\dagger$ indicates the model with the best improvement over the baseline, while $^*$ indicates the model with the best overall performance.}
\label{tab:ood_results}
\end{table}
   
\paragraph{Finetuning, KD and claim normalization all improve robustness for weaker embedding models and can be applied independently or jointly.} KD and finetuning are \textit{train-time} interventions that update model weights to better handle differences between unperturbed and perturbed input claims. KD relies on synthetically generated perturbed input claims that are scalable and inexpensive to produce \cite{shliselberg2024syndy, liu2024best}. In contrast, finetuning requires \textit{claim-factcheck} pairs, which are costly to annotate. On the other hand, claim normalization is an \textit{inference-time} intervention that leverages the innate representational capabilities of weaker embedding models by standardizing perturbed input claims into a form that is easier to represent, without having to update model weights. Although claim normalization is appealing for its ``train-free,'' inference-time application, its reliance on a strong normalizer (often an LLM) may add a computational overhead to the overall retrieval pipeline, making robust adaptation through knowledge distillation a scalable alternative for scenarios where train-time interventions are feasible.

We expect a significant drop in retrieval performance across all approaches for novel claims in the OOD dataset as they are (a) taken from a different time period, (b) cover different topics, and (c) come from a different platform, with different characteristics than the tweets in \textit{CheckThat22} and \textit{FactCheckTweet} (e.g., longer length). We observe such a drop. Nonetheless, \textbf{all mitigation approaches improve performance relative to the baseline and combining them yields the largest improvement of 10 percentage points} similar to the in-domain setting. This highlights the effectiveness of the mitigation approaches in generalizing to novel out-of-domain claims.

\section{Discussion and Conclusion}
Real-world claim matching systems must handle not only factual but also edited claims, which often arise as users engage with misinformation online. Embedding models used in these systems must therefore be robust to such user-informed misinformation edits. However, standard benchmarks like MTEB~\cite{muennighoff2022mteb, enevoldsen2025mmteb}, primarily evaluate sentence embedding models on clean, generalist tasks and fail to capture this challenge. To address this gap, we introduce a perturbation generation framework that produces natural and valid claim edits. This framework creates a more realistic evaluation test bed for sentence embedding models. Although developed for claim matching, these perturbations have broader relevance and can help select embedding models for other downstream tasks that involve user-generated inputs. Our evaluation shows that widely-used embedding models are highly sensitive to user edits, while LLM-distilled embeddings offer greater robustness. Reranking helps, but even strong rerankers cannot fully recover performance under perturbation. To address the retrieval gaps, we show that train- and inference-time approaches can improve embedding model robustness and generalize to out-of-domain settings with different trade-offs. Knowledge distillation and finetuning are train-time interventions that require generating synthetic data or collecting costly human annotated data with additional computational costs for training and re-embedding fact-checks. On the other hand, claim normalization is a ``train-free'' inference-time mitigation that depends on an LLM and may introduce significant latency and cost to the overall pipeline. Encouragingly, prior work~\cite{ma-etal-2023-query} shows that rewriting capabilities can be distilled into smaller models, offering a practical compromise. Overall, our study offers practical insights for improving claim matching systems, ultimately empowering human fact-checkers to retrieve relevant fact-checks with greater reliability.

\section*{Limitations}  
While our study makes meaningful strides in evaluating the robustness of embedding models to real-world user edits and exploring mitigation strategies, we acknowledge some limitations:

\paragraph{Dataset Coverage} Our evaluation relies on claim matching datasets from fact-checking organizations. Although these datasets provide a useful starting point, they represent only a fraction of the broader misinformation landscape, as many claims in the wild go unchecked. As such, the taxonomy of common edits proposed in this work is representative but not exhaustive, as it is limited to claims encountered by fact-checkers. This limitation is evident in our out-of-domain evaluations, where models experience notable performance degradation when faced with novel claims spanning different timelines, topics, and platforms. Nonetheless, our proposed interventions generalize well to the out-of-domain setting, improving the performance of baseline models by an average of 10 percentage points.

\paragraph{Language Scope} Our analysis is limited to English-language claims, even though misinformation is a multilingual phenomenon \cite{kazemi-etal-2021-claim, quelle2025lost}. We believe that our approach could be extended to other languages, and pursuing this extension is an important direction for future work.

\paragraph{Experimental Coverage} Although our experiments cover a diverse range of settings, evaluating $12$ embedding models, $7$ rerankers, and $14$ perturbation types across various MAP@$k$ values ($k \in \{5, 10, 20, 50\}$) and different top-$j$ reranking candidates ($j \in \{5, 10, 20, 50, 100, 200\}$)---this selection is not exhaustive. Alternative configurations may yield different robustness outcomes. We evaluate widely used embedding models that represent various pre-training approaches, backbone architectures and embedding dimension sizes. However, we note that other types of embedding models, such as unsupervised embeddings~\cite{gao-etal-2021-simcse} or contextual document embeddings~\cite{morris2025contextual}, might behave differently. We also do not evaluate closed-source embedding models (e.g., \texttt{text-embedding-ada-002} from OpenAI\footnote{\url{https://platform.openai.com/docs/guides/embeddings}} or \texttt{Embed v3} from Cohere\footnote{\url{https://cohere.com/blog/introducing-embed-v3}}) due to the inference costs involved. While we test different rerankers and select the best-performing one, we acknowledge that the reranker we select could also exhibit some brittleness to the perturbations, and we believe that evaluating rerankers against the same perturbation set would be a valuable direction for future work.

\section*{Ethical Considerations}
We acknowledge the ethical responsibilities inherent in misinformation research given its real-world impact. Our perturbation generation framework relies on widely available LLMs, and our prompts are publicly released for reproducibility; however, this introduces the risk of misuse, whether inadvertently by non-experts or deliberately by malicious actors to generate misinformation or create rewritings that real-world claim matching systems struggle to identify. For a more detailed discussion on the ethical risks associated with LLM-generated misinformation, see \citet{chen2024llmgenerated}. We also note that the synthetically perturbed claims generated by our work may contain fictional or inaccurate information; as such, these claims should not be used to train models that learn facts from data. Notably, during our claim normalization experiments, \texttt{GPT-4o} refused to process two claims containing sensitive political or religious statements. This incident highlights the critical need for automated systems to differentiate between the use of misinformation (deliberately propagating false information) and the mention of misinformation (discussing or referencing sensitive topics without endorsing them) \cite{gligoric-etal-2024-nlp}.

\section*{Acknowledgments}
We thank Meedan for providing the datasets used in this project. JM is supported by the Rhodes Scholarship. SH is supported by a grant from The Alan Turing Institute for the project entitled ``Effective discovery, tracking, and response to mis- and disinformation'' and the ESRC Digital Good Network (grant reference ES/X502352/1). We are grateful to Samuel Mensah and Elena Kochkina for their helpful feedback on drafts of this paper.

\bibliography{anthology,custom}

\clearpage 

\appendix
\section{Dataset Statistics}
\label{appendix:dataset_statistics}
We report the details of each dataset used in our experiments in this section. The \textit{CheckThat22} dataset~\cite{nakov2021clef} is available under a permissible research-use license.\footnote{\url{https://gitlab.com/checkthat_lab/clef2022-checkthat-lab/clef2022-checkthat-lab.git}} Similarly, the \textit{FactCheckTweet} dataset~\cite{kazemi2022matching} can be accessed for research purposes.\footnote{\url{https://lit.eecs.umich.edu/publications.html}} The \textit{OOD Dataset} is provided by Meedan, primarily sourced through five fact-checking organizations operating tiplines on WhatsApp, and our use is consistent with its intended use.
\subsection{Dataset Splits}
For all experiments, we use only the English subsets of the datasets. In the robustness experiments, we apply perturbations to the test set only. \textit{CheckThat22} provides a predefined train/dev/test split, whereas \textit{FactCheckTweet} does not. We therefore partition the latter into 80\% training, 10\% development, and 10\% test splits. The train and development splits from \textit{CheckThat22} are used for finetuning the embedding models or improving robustness through the teacher-student knowledge distillation approach. For the \textit{OOD Dataset}, we treat the full split as an evaluation set and do not partition it. Table~\ref{tab:dataset_splits} shows the splits for each dataset.

\begin{table}[ht]
    \centering
    \begin{tabular}{lccc}
        \toprule
        \textbf{Dataset} & \textbf{Train} & \textbf{Dev} & \textbf{Test} \\
        \midrule
        \textit{CheckThat22} & 1,195 & 1,195 & 202 \\
        \textit{FactCheckTweet} & 2,504 & 537 & 537 \\
        \textit{OOD Dataset} & - & - & 721 \\
        \bottomrule
    \end{tabular}
    \caption{Dataset splits for training, development, and testing.}
    \label{tab:dataset_splits}
\end{table}

\subsection{Perturbed Claims Statistics}
We report the number of valid perturbed input claims for each perturbation type in Table~\ref{tab:valid_perturbated_claims_statistics}. All perturbations are applied using the framework detailed in Section~\ref{sec:generationframework}, except for Casing, which we apply using the \texttt{TrueCase} Python library.\footnote{\url{https://github.com/daltonfury42/truecase}} While perturbations are initially applied to the full test sets, the number of valid claims varies due to differences in the selection criteria introduced by the LLM as a verifier. \textbf{When comparing against unperturbed claims, we use only the same subset and not the entire original test set.} Overall, perturbations applied at the maximum noise level (\(b_{\text{high}}\)) result in more invalid perturbations, which are filtered out in our generation pipeline.
\begin{table}[ht]
    \centering
    \resizebox{1.0\textwidth}{!}{%
    \begin{tabular}{llcc}
        \toprule
        \textbf{Perturbation} & \textbf{Type} & \textit{\textbf{CheckThat22}} & \textit{\textbf{FactCheckTweet}} \\
        \midrule
        \multirow{2}{*}{Casing} 
        & \textit{TrueCase} & 202 & 537 \\
        & \textit{UpperCase} & 202 & 537 \\
        \midrule
        \multirow{2}{*}{Typos} 
        & \textit{Least} & 192 & 443 \\
        & \textit{Most} & 189 & 441 \\
        \midrule
        \multirow{2}{*}{Negation} 
        & \textit{Shallow} & 181 & 405 \\
        & \textit{Double} & 148 & 308 \\
        \midrule
        \multirow{2}{*}{Entity Rep.} 
        & \textit{At least 1} & 181 & 327 \\
        & \textit{All} & 110 & 164 \\
        \midrule
        \multirow{2}{*}{LLM Rewrite} 
        & \textit{Least} & 185 & 373 \\
        & \textit{Most} & 185 & 371 \\
        \midrule
        \multirow{4}{*}{Dialect} 
        & \textit{AAE} & 195 & 449 \\
        & \textit{Jamaican} & 192 & 443 \\
        & \textit{Pidgin} & 192 & 449 \\
        & \textit{Singlish} & 195 & 449 \\
        \bottomrule
    \end{tabular}
    }
    \caption{Valid perturbed input claims for each perturbation type across \textit{CheckThat22} and \textit{FactCheckTweet} datasets.}
    \label{tab:valid_perturbated_claims_statistics}
\end{table}

We report the degree of lexical overlap between the unperturbed and perturbed input claims for the \textit{CheckThat22} and \textit{FactCheckTweet} datasets in Table~\ref{tab:check_that_perturbation_analysis} and Table~\ref{tab:fact_check_perturbation_analysis}, respectively. Following the approach of \citet{chiang-etal-2023-revealing}, we compute the ROUGE F1 scores (\textbf{R1}, \textbf{R2}, \textbf{RL}) \cite{lin2004rouge}, where R1 measures unigrams, R2 measures bigrams, and RL measures the longest common subsequence. We use the \texttt{rouge}\footnote{\url{https://pypi.org/project/rouge/}} package for this purpose. A higher ROUGE score indicates greater lexical similarity. We also report the normalized Levenshtein distance between the unperturbed and perturbed input pairs, obtained by calculating the Levenshtein distance between the pair and dividing by the length of the longer sequence. We tokenize the sentences using \texttt{bert-base-uncased} and calculate the distance between the token IDs. A lower Levenshtein distance implies higher lexical overlap. Across all perturbation types, we observe that the worst‐case edits---those applied with the maximum noise level (\(b_{\text{high}}\))---result in perturbed claims that have less lexical overlap with their unperturbed counterparts than do the edits applied with the minimum noise level (\(b_{\text{low}}\)). This difference in lexical overlap enables us to capture a wide spectrum of user edits.

\begin{table}[ht]
\centering
\resizebox{1.0\textwidth}{!}{%
\begin{tabular}{llcccc}
\toprule
\textbf{Perturbation} & \textbf{Type}       & \textbf{R1}  & \textbf{R2}  & \textbf{RL}  & \textbf{Lev.} \\
\midrule
\multirow{2}{*}{Casing} 
             & \textit{TrueCase}  & 90.20      & 83.02      & 90.20      & 0.00 \\
             & \textit{UpperCase} &  0.00      &  0.00      &  0.00      & 0.00 \\
\midrule
\multirow{2}{*}{Typos} 
             & \textit{Least}     & 60.47      & 31.46      & 60.47      & 0.48 \\
             & \textit{Most}      & 41.67      & 10.81      & 38.89      & 0.68 \\
\midrule
\multirow{2}{*}{Negation} 
             & \textit{Shallow}   & 56.34      & 46.58      & 53.52      & 0.65 \\
             & \textit{Double}    & 17.78      & 12.50      & 17.78      & 0.90 \\
\midrule 
\multirow{2}{*}{Entity Rep.} 
             & \textit{Atleast 1} & 98.00      & 94.23      & 98.00      & 0.05 \\
             & \textit{All}       & 87.50      & 74.75      & 87.50      & 0.14 \\
\midrule
\multirow{2}{*}{LLM Rewrite} 
             & \textit{Least}     & 58.14      & 30.77      & 53.49      & 0.52 \\
             & \textit{Most}      & 41.46      & 16.47      & 41.46      & 0.66 \\
\midrule
\multirow{4}{*}{Dialect} 
             & \textit{AAE}                & 39.17      & 17.19      & 37.97      & 0.63 \\
             & \textit{Jamaican}     & 34.33      & 12.98      & 32.98      & 0.68 \\
             & \textit{Pidgin}             & 34.49      & 13.32      & 33.00      & 0.68 \\
             & \textit{Singlish}           & 40.28      & 17.81      & 38.51      & 0.65 \\
\bottomrule
\end{tabular}%
}
\caption{Degree of lexical overlap between unperturbed and perturbed input claims for the \textit{\textbf{CheckThat22}} dataset. The table reports ROUGE F1 scores (R1, R2, RL), and the normalized Levenshtein distance (Lev.). Higher ROUGE scores and lower Levenshtein distances indicate higher lexical overlap.}
\label{tab:check_that_perturbation_analysis}
\end{table}

\begin{table}[ht]
\centering
\resizebox{1.0\textwidth}{!}{
\begin{tabular}{llcccc}
\toprule
\textbf{Perturbation} & \textbf{Type}       & \textbf{R1} & \textbf{R2} & \textbf{RL} & \textbf{Lev.} \\
\midrule
\multirow{2}{*}{Casing} 
             & \textit{TrueCase}  & 81.25     & 66.67     & 81.25     & 0.00 \\
             & \textit{UpperCase} & 6.25      & 0.00      & 6.25      & 0.00 \\
\midrule
\multirow{2}{*}{Typos} 
             & \textit{Least}     & 42.86     & 30.77     & 42.86     & 0.55 \\
             & \textit{Most}      & 42.86     & 15.38     & 35.71     & 0.95 \\
\midrule
\multirow{2}{*}{Negation} 
             & \textit{Shallow}   & 96.77     & 89.66     & 96.77     & 0.11 \\
             & \textit{Double}    & 77.78     & 66.67     & 77.78     & 0.31 \\
\midrule
\multirow{2}{*}{Entity Rep.} 
             & \textit{At least 1} & 93.75     & 93.33     & 93.75     & 0.21 \\
             & \textit{All}       & 46.67     & 14.29     & 46.67     & 0.63 \\
\midrule
\multirow{2}{*}{LLM Rewrite} 
             & \textit{Least}     & 48.48     & 19.35     & 48.48     & 0.58 \\
             & \textit{Most}      & 43.75     & 20.00     & 37.50     & 0.86 \\
\midrule
\multirow{4}{*}{Dialect} 
             & \textit{AAE} & 33.92 & 12.11 & 32.29 & 0.71 \\
             & \textit{Jamaican}          & 29.52 & 9.42  & 27.92 & 0.75 \\
             & \textit{Pidgin}                   & 30.37 & 10.25 & 28.78 & 0.74 \\
             & \textit{Singlish}                 & 35.45 & 14.15 & 33.50 & 0.71 \\
\bottomrule
\end{tabular}
}
\caption{Degree of lexical overlap between unperturbed and perturbed input claims for the \textit{\textbf{FactCheckTweet}} dataset. The table reports ROUGE F1 scores (R1, R2, RL) and the normalized Levenshtein distance (Lev.). Higher ROUGE scores and lower Levenshtein distances indicate higher lexical overlap.}
\label{tab:fact_check_perturbation_analysis}
\end{table}

\section{Human Evaluation}
\label{appendix:human_evaluation}
We conduct a human evaluation to assess the effectiveness of our perturbation generation framework in producing \textit{\textbf{valid}} and \textit{\textbf{natural}} perturbed claims. Our evaluation follows a similar annotation task to that of \citet{dyrmishi_how_2023}, which assessed adversarial examples in human settings. We randomly sample 20 unique perturbed input claims from three perturbation types—\textit{typos}, \textit{entity replacement}, and \textit{dialect}. Half of these claims are labeled as valid (1) and the other half as invalid (0) by the \textit{LLM as a Verifier}. We then present both the unperturbed and perturbed claims, along with the annotation instructions provided below, to three annotators, all of whom are authors. No compensation was paid to the author-annotators.

\begin{tcolorbox}[
    colback=gray!10, 
    colframe=teal!60!green,  %
    title=Annotation Instructions,
    rounded corners
]

\textbf{Your task:} For each perturbed claim, evaluate the following criteria:

\vspace{0.5em}
\textbf{1. Perturbation Accuracy}  
Does the input claim contain the \texttt{perturbation\_type}?  
\\ \checkboxYes \quad \checkboxNo  

\vspace{0.5em}
\textbf{2. Validity and Naturalness}  
\begin{enumerate}[label=\arabic*., left=0pt, labelindent=0pt]
    \item Does the fact-check apply to it, i.e., is the fact-check helpful in verifying the perturbed claim?  
    \checkboxYes \quad \checkboxNo  
    \item Does it convey the same main claim as the unperturbed claim?  
    \checkboxYes \quad \checkboxNo  
    \item Does the perturbed claim feel like something a typical social media user might write?  
    \checkboxYes \quad \checkboxNo  
\end{enumerate}
\end{tcolorbox}
We first assess whether the perturbed claim contains the intended perturbation type compared to the original. We define this as perturbation accuracy, measured as the proportion of instances in which the LLM as a Perturber successfully applies the intended change. Next, we evaluate validity (and naturalness) by collecting binary labels from three independent annotators, using an all-or-nothing rule where a positive label is assigned only if all annotators agree. For validity, we report precision on the positive class, reflecting the reliability of the LLM as a Verifier in identifying valid perturbations. Table~\ref{tab:human_evaluation} shows that our framework achieves a perturbation accuracy of 96.70\%, indicating that the intended perturbation was applied in nearly all cases. The verifier attained a 100\% precision on the positive class, meaning that every claim labelled as valid by the model was confirmed by human annotators. However, the overall recall score for validity is lower (55.67\%), suggesting that while the verifier is highly precise, it may be overly conservative, resulting in some valid instances getting filtered out. 

\begin{table}[ht]
    \centering
    \resizebox{\columnwidth}{!}{%
    \begin{tabular}{lccccc}
        \toprule
        \textbf{Type} & \textbf{\# Count} & \textbf{Acc (\%)} & \textbf{Valid (Pr.)} & \textbf{Valid (Re.)} & \textbf{Valid (F1)} \\
        \midrule
        \textit{Typos}         & $20$ & $100$  & $100$ & $52.63$ & $68.97$ \\
        \textit{Entity Rep.}   & $20$ & $90$   & $100$ & $58.82$ & $72.07$ \\
        \textit{Dialect}       & $20$ & $100$  & $100$ & $55.56$ & $71.43$ \\
        \midrule
        \textbf{Total}         & $60$ & $96.70$& $100$ & $55.67$ & $70.82$ \\
        \bottomrule
    \end{tabular}%
    }
    \caption{\textbf{Perturbation Evaluation:} \textbf{Acc (\%)} indicates the proportion of instances where the perturbation was correctly applied by the \textit{LLM As a Perturber}. \textbf{Valid (Pr.)} represents the precision on valid instances, \textbf{Valid (Re.)} is the recall on valid instances, and \textbf{Valid (F1)} is the overall F1 score.}
    \label{tab:human_evaluation}
\end{table}

\section{Embedding Models Details}
\label{appendix:embedding_models}
The list of all the embedding models evaluated in this study; their parameter count and embedding dimensions are shown in Table \ref{tab:embedding_models}. We use all embedding models off-the-shelf with default hyperparameters, utilizing the \texttt{SentenceTransformers} library.\footnote{\url{https://sbert.net/}} Since the \textit{FactCheckTweet} dataset contains longer articles and most embedding models support only up to \(512\) tokens, we split articles into paragraphs and compute similarity between each paragraph and the input text, following \citet{kazemi-etal-2021-claim}. This approach is also applied to LLM-based embeddings, despite their support for longer sequences, to ensure fair comparison. To run the BM25 evaluation, we used the \texttt{rank\_bm25} implementation.\footnote{\url{https://github.com/dorianbrown/rank_bm25}}

\begin{table}[ht]
    \centering
    \resizebox{1.0\textwidth}{!}{%
    \begin{tabular}{lcc}
        \toprule
        \textbf{Embedding Model} & \textbf{\# Params} & \textbf{$d_{emb}$}  \\
        \midrule
        \texttt{\href{http://huggingface.co/sentence-transformers/all-MiniLM-L12-v2}{all-MiniLM-L12-v2}} & 82M & 384 \\
        \texttt{\href{https://huggingface.co/sentence-transformers/all-distilroberta-v1}{all-distilroberta-v1}} & 82M & 768 \\
        \texttt{\href{https://huggingface.co/sentence-transformers/all-mpnet-base-v2}{all-mpnet-base-v2}} & 110M & 768 \\
        \texttt{\href{https://huggingface.co/sentence-transformers/sentence-t5-base}{sentence-t5-base}} & 110M & 768 \\
        \texttt{\href{https://huggingface.co/sentence-transformers/sentence-t5-large}{sentence-t5-large}} & 335M & 768 \\
        \texttt{\href{https://huggingface.co/hkunlp/instructor-base}{hkunlp/instructor-base}} & 86M & 768 \\
        \texttt{\href{https://huggingface.co/hkunlp/instructor-large}{hkunlp/instructor-large}} & 335M & 768 \\
        \texttt{\href{https://huggingface.co/nvidia/NV-Embed-v2}{nvidia/NV-Embed-v2}} & 7B & 4,096 \\
        \texttt{\href{https://huggingface.co/Salesforce/SFR-Embedding-Mistral}{Salesforce/SFR-Embedding-Mistral}} & 7B & 4,096 \\
        \bottomrule
    \end{tabular}
    \caption{Embedding models with their parameter counts (\# Params) and embedding dimensions ($d_{emb}$). Model names are hyperlinked.}
    \label{tab:embedding_models}
    }
\end{table}

\subsection{Training Setup and Hyperparameters}
\label{appendix:training_setup}
\begin{table*}[ht]
\centering
\resizebox{\textwidth}{!}{%
\begin{tabular}{p{0.32\textwidth} p{0.32\textwidth} p{0.25\textwidth}}
\toprule
\textbf{Source Sentence ($s_i$)}& \textbf{Target Sentence ($t_i$)}& \textbf{Pairing ($s_i$ vs $t_i$)} \\
\midrule
The US drone attack on \#Soleimani caught on camera. \#IranUsa & Dem catch US drone strike on Soleimani pon camera, yah. \#IranUsa & \textit{Unperturbed} vs.\ \textit{Jamaican Patois}\\
\midrule
The American drone strike on \#Qassem caught on camera. \#TehranWashington & Dem capture US drone attack wey hit Soleimani for cam. \#IranUsa & \textit{Entity Replacement} vs.\ \textit{Nigerian Pidgin}\\
\midrule
It is not false that the US drone attack on \#Soleimani was caught on camera. & Wah, US drone attack on Soleimani kenna caught on video leh. \#IranUsa & \textit{Negation} vs.\ \textit{Singlish} \\
\midrule
US drone strike on Soleimani captrd on video. \#IranUsa. & US drone strike pon Soleimani seen on di camera, seen. \#IranUsa & \textit{Typos} vs.\ \textit{Jamaican Patois} \\
\bottomrule
\end{tabular}%
}
\caption{Examples of parallel sentences generated using various perturbations.}
\label{tab:parallel_sentences}
\end{table*}

\subsubsection{Finetuning Embedding Models}
\label{appendix:finetuning}
We investigate the robustness of finetuned embedding models by finetuning \texttt{all-mpnet-base-v2} and \texttt{sentence-t5-base} on the \textit{CheckThat22} training split, which consists of \(1,195\) input claims. We follow a similar approach to \citet{shliselberg2022riet} and finetune the embedding models using contrastive learning with the Multiple Negatives Ranking (MNR) loss. MNR loss optimizes the similarity between positive claim–factcheck pairs while reducing the similarity of negative pairs within a batch. It is defined as:

\begin{equation}
- \frac{1}{n} \sum_{i=1}^{n} \log \frac{\exp(\text{sim}(a_i, p_i))}{\sum_{j} \exp(\text{sim}(a_i, p_j))}
\end{equation}

where \( f(x) \) and \( f(y) \) are the embedding representations of \( x \) (input claim) and \( y \) (factcheck), respectively. To improve generalization, we introduce ``hard negatives'' by using BM25 to generate a ranked list of candidate negatives and selecting the top-ranked negative factcheck. We use the training scripts provided by \citet{shliselberg2022riet}\footnote{\url{https://github.com/RIET-lab/GenerativeClaimMatchingPipeline/blob/main/src/dynamicquery/candidate_selection/train_sentence_model.py}} with the default hyperparameters: \texttt{learning rate} = \( 5e-6\), \texttt{batch size} = \( 6 \), \texttt{max length} = \( 128 \), \texttt{temperature} = \( 0.1 \), and \texttt{epochs} = \( 1 \).

\subsubsection{Teacher-Student Knowledge Distillation}
\label{appendix:teacher_student}
We provide the details of the knowledge-distillation approach for improving the robustness of weaker embedding models in this section. This approach depends on parallel sentence pairs $\bigl((s_1, t_1), \dots, (s_n, t_n)\bigr)$ where $s_i$ is a source sentence and $t_i$ the target sentence. In our case,  $s_i$ and $t_i$ are variations of the same claim. We obtain these parallel sentences by generating perturbations of the \textit{CheckThat22} training split for \textit{typos}, \textit{negation}, \textit{dialect}, and \textit{entity replacement}, resulting in an initial set of \(11,593\) pairs. We then expand this list by pairing instances of the same claim (e.g., \textit{dialect} vs.\ \textit{typos}), yielding a larger dataset of \(70,954\) pairs. Examples of parallel sentences are shown in Table ~\ref{tab:parallel_sentences}. We adopt the same training approach as in \citet{reimers-gurevych-2020-making} where we train a student model $\hat{M}$ such that 
\[
\hat{M}(s_i) \approx M(s_i) \quad \text{and} \quad \hat{M}(t_i) \approx M(s_i).
\]
For a given minibatch $B$, we minimize the mean-squared loss, \begin{align*}
\mathcal{L}_{\text{MSE}} &= \frac{1}{|B|} \sum_{j \in B} \Biggl[
  \left( M(s_j) - \hat{M}(s_j) \right)^2 \\
  &\quad + \left( M(s_j) - \hat{M}(t_j) \right)^2
\Biggr]
\end{align*}
We use \texttt{all-mpnet-base-v2} as both the student ($\hat{M}$) and teacher (${M}$) model. To train the model, we use the script provided by \citet{reimers-gurevych-2020-making}\footnote{\url{https://github.com/UKPLab/sentence-transformers/blob/master/examples/training/multilingual/make_multilingual.py}} with the following hyperparameters: \texttt{learning rate} = \( 2e-5\), \texttt{batch size} = \( 64 \), \texttt{max length} = \( 256 \), and \texttt{epochs} = \( 20 \).

\section{Rerankers Evaluation}
\subsection{Selecting a strong reranker}
\label{appendix:selecting_strong_reranker}
Given the many possible combinations between embedding models and rerankers, we evaluate a wide range of reranker models to select a strong-performing one for further evaluation. In our study, we consider several types of rerankers: \textbf{pairwise rerankers}, such as \texttt{ms-marco-MiniLM-L-6-v2} and \texttt{monot5-3b}~\cite{nogueira-etal-2020-document}; \textbf{listwise rerankers}, such as \texttt{LiT5-Distill}~\cite{tamber2023scaling}; and \textbf{LLM-based rerankers}, including \texttt{bge-reranker-v2-gemma}~\cite{chen2024bge}, \texttt{rank\_zephyr\_7b\_v1\_full}~\cite{pradeep2023rankzephyr}, and \texttt{RankGPT}~\cite{Sun2023IsCG} (for which we use \texttt{GPT4o} as the reranker). We evaluate these rerankers using the first-stage retrieval rankings from \texttt{all-mpnet-base-v2} on the unperturbed \textit{CheckThat22} test set, setting the number of candidates, i.e., \textit{top-j}, to 20. A value of 20 is chosen as it is a feasible size that fits well within the context window for \texttt{RankGPT}. We use \texttt{RankLLM}\footnote{\url{https://github.com/castorini/rank_llm}} implementation for the different reranker models. 

Table~\ref{tab:reranker_evaluation} presents the performance results for the different rerankers. RankGPT, using GPT-4o as the reranker, achieves the best performance. However, its results are comparable to the \texttt{bge-reranker-v2-gemma} model, based on \texttt{gemma-2b}. Given that \texttt{bge-reranker-v2-gemma} is lightweight and can be run locally with lower latency and cost, we select it as a representative SoTA reranker over GPT-4o, which increases latency in production retrieval systems~\cite{Sun2023IsCG}.

\begin{table}[H]
    \centering
    \resizebox{\textwidth}{!}{%
        \begin{tabular}{lcccc}
            \toprule
            \textbf{Model} & \textbf{MAP@1} & \textbf{MAP@5} & \textbf{MAP@10} & \textbf{MAP@20} \\
            \midrule
            \texttt{RankGPT(GPT4o)} & \textbf{0.908} & \textbf{0.919} & \textbf{0.919} & \textbf{0.919} \\
            \texttt{\href{https://huggingface.co/BAAI/bge-reranker-v2-gemma}{bge-reranker-v2-gemma}} & 0.908 & 0.918 & 0.918 & 0.918 \\
            \texttt{\href{https://huggingface.co/castorini/monot5-3b-msmarco-10k}{monot5-3b-msmarco-10k}} & 0.897 & 0.912 & 0.912 & 0.912 \\
            \texttt{\href{https://huggingface.co/castorini/LiT5-Distill-large-v2}{LiT5-Distill-large-v2}} & 0.876 & 0.902 & 0.902 & 0.902 \\
            \texttt{\href{https://huggingface.co/BAAI/bge-reranker-v2-m3}{bge-reranker-v2-m3}} & 0.870 & 0.891 & 0.892 & 0.892 \\
            \texttt{\href{https://huggingface.co/cross-encoder/ms-marco-MiniLM-L-6-v2}{ms-marco-MiniLM-L-6-v2}} & 0.854 & 0.883 & 0.884 & 0.884 \\
            \texttt{\href{https://huggingface.co/castorini/rank_zephyr_7b_v1_full}{rank\_zephyr\_7b\_v1\_full}} & 0.778 & 0.845 & 0.846 & 0.846 \\
            \bottomrule
        \end{tabular}%
    }
    \caption{Performance comparison of rerankers on \textit{CheckThat22} (unperturbed test set). The table reports Mean Average Precision (MAP@k) at different cutoffs (\( k = 1, 5, 10, 20 \)) for rerankers applied to first-stage top-20 candidates from \texttt{all-mpnet-base-v2}.}
    \label{tab:reranker_evaluation}
\end{table}

\subsection{Selecting the optimal \textit{top-j} candidates for reranking}
\label{appendix:selecting_optimal_k}
\begin{figure}[ht]
    \centering
    \includegraphics[width=0.9\textwidth]{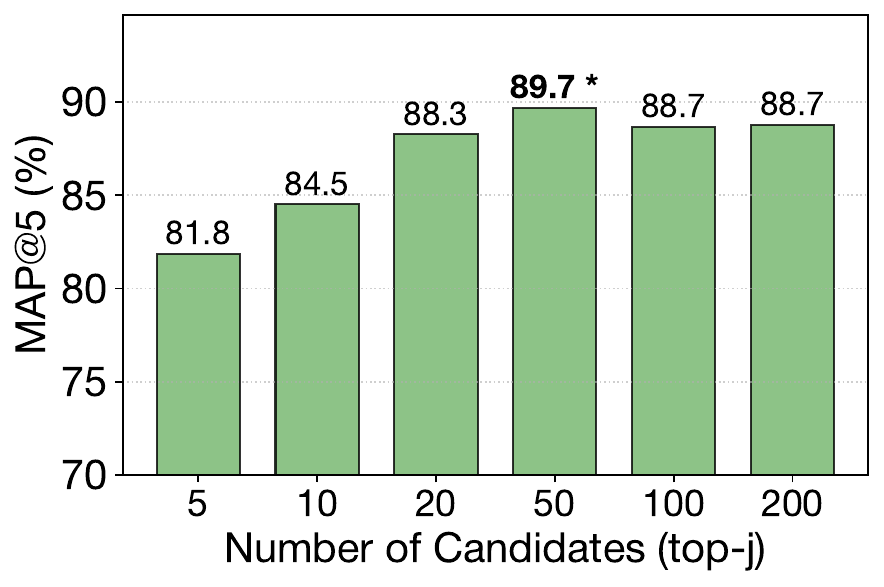}
    \vspace{-2mm}
    \caption{MAP@$5$ performance across different \textit{top-j} values for \textbf{cross-encoder/ms-marco-MiniLM-L-6-v2} on \textit{CheckThat22}. The highest MAP@$5$ is achieved at \( j=50 \).}
    \label{fig:optimal_k_cross_encoder}
\end{figure}

\begin{figure}[ht]
    \centering
    \includegraphics[width=0.9\textwidth]{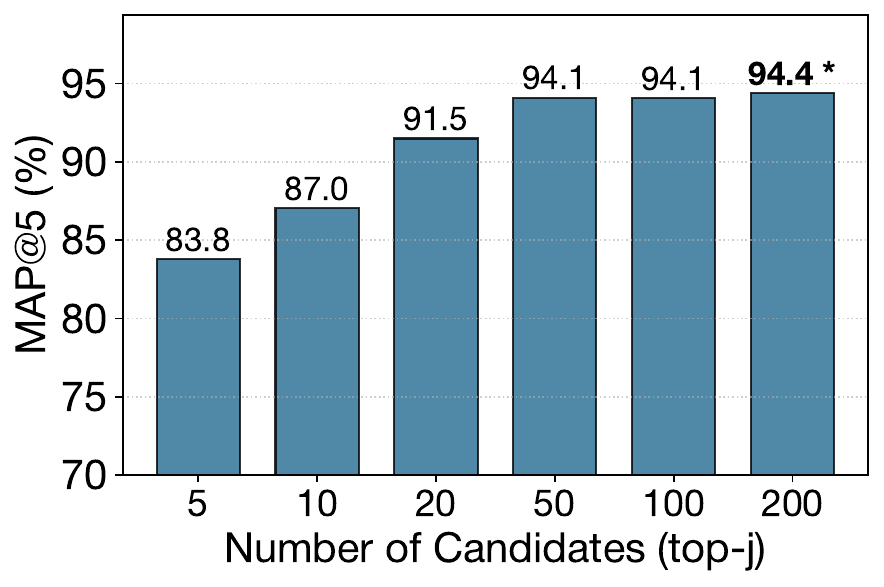}
    \vspace{-2mm}
    \caption{MAP@$5$ performance across different \textit{top-j} values for \textbf{bge-reranker-v2-gemma} on \textit{CheckThat22}. Reranking improves up to \( j=50 \), after which further gains are minimal.}
    \label{fig:optimal_k_bge}
\end{figure}

To determine the optimal number of candidates for reranking (\textit{top-j}), we evaluate MAP@$5$ across different values of \( j \in \{5, 10, 20, 50, 100, 200\} \), on the \textit{CheckThat22} unperturbed test set. We use MAP@$5$ to ensure comparability across different values of \textit{j}. We use two rerankers: \texttt{cross-encoder/ms-marco-MiniLM-L-6-v2} and \texttt{bge-reranker-v2-gemma}, applied to first-stage retrieval rankings from \texttt{all-mpnet-base-v2}. Figure~\ref{fig:optimal_k_cross_encoder} presents the reranking performance of cross-encoder/ms-marco-MiniLM-L-6-v2, while Figure~\ref{fig:optimal_k_bge} shows that of bge-reranker-v2-gemma. The cross-encoder/ms-marco-MiniLM-L-6-v2 achieves its highest MAP@$5$ at \( j=50 \), whereas the bge-reranker-v2-gemma shows minimal improvement beyond this point. Based on this, we set \textit{top-j} to \( 50 \) for the reranking experiments, balancing performance and efficiency.

\section{Compute Requirements}
All embedding model inferences, reranker fine-tuning, and evaluations are conducted on a single NVIDIA A100 80GB GPU. 

\section{Additional Results}
\label{appendix:additional_results}
\subsection{Additional Results Across Different MAP@\textit{k} on \textit{CheckThat22}}
\label{appendix:additional_checkthat_k}
We provide robustness evaluation results across different MAP@\textit{k} values, \( k \in \{5, 10, 50\} \), for the \textit{CheckThat22} dataset. Table~\ref{tab:map@5_checkthat22} reports the evaluation at MAP@\textit{5}, Table~\ref{tab:map@10_checkthat22} at MAP@\textit{10}, and Table~\ref{tab:map@50_checkthat22} at MAP@\textit{50}, with reranking applied to the top 50 candidates in all cases. We observe minimal performance variations across different MAP@\textit{k} values on the \textit{CheckThat22} dataset, indicating that the robustness results hold across various MAP@\textit{k} values.
\begin{table*}[t!]
\centering
\resizebox{0.9\textwidth}{!}{%
\begin{tabular}{ll|cc|cc|cc|cc|cc|cccc}
\toprule
 \multicolumn{2}{l}{} & \multicolumn{2}{c|}{\textbf{Casing}} & \multicolumn{2}{c|}{\textbf{Typos}} & \multicolumn{2}{c|}{\textbf{Negation}} & \multicolumn{2}{c|}{\textbf{Entity Replacement}} & \multicolumn{2}{c|}{\textbf{LLM Rewrite}} & \multicolumn{4}{c}{\textbf{Dialect}} \\
 \textbf{Stage} & \textbf{Model} & \textit{TrueCase} & \textit{Upper} & \textit{Least} & \textit{Most} & \textit{Shallow} & \textit{Double} & \textit{Atleast 1} & \textit{All} & \textit{Least} & \textit{Most} & \textit{AAE} & \textit{Jamaican} & \textit{Pidgin} & \textit{Singlish} \\
\midrule
\parbox[t]{2mm}{\multirow{12}{*}{\rotatebox[origin=c]{90}{
 \parbox[c]{2cm}{\centering  \textbf{First-Stage Retrieval}}}}} 
&\texttt{BM25} & \midpointgradientcell{-0.06}{-35.92}{5.44}{0}{neg}{pos}{\opacity}{0} & \midpointgradientcell{0.00}{-35.92}{5.44}{0}{neg}{pos}{\opacity}{0} & \midpointgradientcell{-15.33}{-35.92}{5.44}{0}{neg}{pos}{\opacity}{0} & \midpointgradientcell{-15.19}{-35.92}{5.44}{0}{neg}{pos}{\opacity}{0} & \midpointgradientcell{0.57}{-35.92}{5.44}{0}{neg}{pos}{\opacity}{0} & \midpointgradientcell{-2.59}{-35.92}{5.44}{0}{neg}{pos}{\opacity}{0} & \midpointgradientcell{-2.79}{-35.92}{5.44}{0}{neg}{pos}{\opacity}{0} & \midpointgradientcell{-12.83}{-35.92}{5.44}{0}{neg}{pos}{\opacity}{0} & \midpointgradientcell{-0.85}{-35.92}{5.44}{0}{neg}{pos}{\opacity}{0} & \midpointgradientcell{-0.25}{-35.92}{5.44}{0}{neg}{pos}{\opacity}{0} & \midpointgradientcell{-9.18}{-35.92}{5.44}{0}{neg}{pos}{\opacity}{0} & \midpointgradientcell{-6.64}{-35.92}{5.44}{0}{neg}{pos}{\opacity}{0} & \midpointgradientcell{-5.03}{-35.92}{5.44}{0}{neg}{pos}{\opacity}{0} & \midpointgradientcell{-1.71}{-35.92}{5.44}{0}{neg}{pos}{\opacity}{0} \\
\bottomrule
\end{tabular}}
\caption{Effect of perturbations on \textbf{\textit{CheckThatTweet}} (\(\text{MAP\textit{@}}5\)). \textbf{First-Stage Retrieval} shows the \emph{retrieval gap} between unperturbed and perturbed inputs: $\Delta(\text{MAP}@5)_{\text{retrieval}}$. \textbf{Reranking Recovery} measures the \(\Delta \text{\emph{recovery gap}}\) for the perturbed input set before and after reranking top-\textit{50} candidates, with improvements highlighted in \hlrerank{green} and drops in \hlneg{red}. \textbf{Overall Pipeline} shows the \(\Delta \text{\emph{overall gap}}\) for combined retrieval and reranking between unperturbed and perturbed inputs. Colors indicate \hlpos{positive} or \hlneg{negative} deltas compared to unperturbed. \(\text{MAP\textit{@}}5\) is used for all comparisons. Models finetuned on the task are indicated with the postfix \texttt{ft}.\vspace{-0.5em}}
\label{tab:map@5_checkthat22}
\end{table*}

\begin{table*}[t!]
\centering
\resizebox{0.9\textwidth}{!}{%
\begin{tabular}{ll|cc|cc|cc|cc|cc|cccc}
\toprule
 \multicolumn{2}{l}{} & \multicolumn{2}{c|}{\textbf{Casing}} & \multicolumn{2}{c|}{\textbf{Typos}} & \multicolumn{2}{c|}{\textbf{Negation}} & \multicolumn{2}{c|}{\textbf{Entity Replacement}} & \multicolumn{2}{c|}{\textbf{LLM Rewrite}} & \multicolumn{4}{c}{\textbf{Dialect}} \\
 \textbf{Stage} & \textbf{Model} & \textit{TrueCase} & \textit{Upper} & \textit{Least} & \textit{Most} & \textit{Shallow} & \textit{Double} & \textit{Atleast 1} & \textit{All} & \textit{Least} & \textit{Most} & \textit{AAE} & \textit{Jamaican} & \textit{Pidgin} & \textit{Singlish} \\
\midrule
\parbox[t]{2mm}{\multirow{12}{*}{\rotatebox[origin=c]{90}{
 \parbox[c]{2cm}{\centering  \textbf{First-Stage Retrieval}}}}} 
&\texttt{BM25} & \midpointgradientcell{0.01}{-35.57}{5.68}{0}{neg}{pos}{\opacity}{0} & \midpointgradientcell{0.00}{-35.57}{5.68}{0}{neg}{pos}{\opacity}{0} & \midpointgradientcell{-15.32}{-35.57}{5.68}{0}{neg}{pos}{\opacity}{0} & \midpointgradientcell{-15.19}{-35.57}{5.68}{0}{neg}{pos}{\opacity}{0} & \midpointgradientcell{0.42}{-35.57}{5.68}{0}{neg}{pos}{\opacity}{0} & \midpointgradientcell{-2.62}{-35.57}{5.68}{0}{neg}{pos}{\opacity}{0} & \midpointgradientcell{-2.81}{-35.57}{5.68}{0}{neg}{pos}{\opacity}{0} & \midpointgradientcell{-12.48}{-35.57}{5.68}{0}{neg}{pos}{\opacity}{0} & \midpointgradientcell{-0.82}{-35.57}{5.68}{0}{neg}{pos}{\opacity}{0} & \midpointgradientcell{-0.32}{-35.57}{5.68}{0}{neg}{pos}{\opacity}{0} & \midpointgradientcell{-9.28}{-35.57}{5.68}{0}{neg}{pos}{\opacity}{0} & \midpointgradientcell{-6.84}{-35.57}{5.68}{0}{neg}{pos}{\opacity}{0} & \midpointgradientcell{-5.17}{-35.57}{5.68}{0}{neg}{pos}{\opacity}{0} & \midpointgradientcell{-1.75}{-35.57}{5.68}{0}{neg}{pos}{\opacity}{0} \\
\bottomrule
\end{tabular}}
\caption{Effect of perturbations on \textbf{\textit{CheckThat22}} (\(\text{MAP\textit{@}}10\)). \textbf{First-Stage Retrieval} shows the \emph{retrieval gap} between unperturbed and perturbed inputs: $\Delta(\text{MAP}@10)_{\text{retrieval}}$. \textbf{Reranking Recovery} measures the \(\Delta \text{\emph{recovery gap}}\) for the perturbed input set before and after reranking top-\textit{50} candidates, with improvements highlighted in \hlrerank{green} and drops in \hlneg{red}. \textbf{Overall Pipeline} shows the \(\Delta \text{\emph{overall gap}}\) for combined retrieval and reranking between unperturbed and perturbed inputs. Colors indicate \hlpos{positive} or \hlneg{negative} deltas compared to unperturbed. \(\text{MAP\textit{@}}10\) is used for all comparisons. Models finetuned on the task are indicated with the postfix \texttt{ft}.\vspace{-0.5em}}
\label{tab:map@10_checkthat22}
\end{table*}

\begin{table*}[t!]
\centering
\resizebox{0.9\textwidth}{!}{%
\begin{tabular}{ll|cc|cc|cc|cc|cc|cccc}
\toprule
 \multicolumn{2}{l}{} & \multicolumn{2}{c|}{\textbf{Casing}} & \multicolumn{2}{c|}{\textbf{Typos}} & \multicolumn{2}{c|}{\textbf{Negation}} & \multicolumn{2}{c|}{\textbf{Entity Replacement}} & \multicolumn{2}{c|}{\textbf{LLM Rewrite}} & \multicolumn{4}{c}{\textbf{Dialect}} \\
 \textbf{Stage} & \textbf{Model} & \textit{TrueCase} & \textit{Upper} & \textit{Least} & \textit{Most} & \textit{Shallow} & \textit{Double} & \textit{Atleast 1} & \textit{All} & \textit{Least} & \textit{Most} & \textit{AAE} & \textit{Jamaican} & \textit{Pidgin} & \textit{Singlish} \\
\midrule
\parbox[t]{2mm}{\multirow{12}{*}{\rotatebox[origin=c]{90}{
 \parbox[c]{2cm}{\centering  \textbf{First-Stage Retrieval}}}}}
&\texttt{BM25} & \midpointgradientcell{0.01}{-35.26}{5.29}{0}{neg}{pos}{\opacity}{0} & \midpointgradientcell{0.00}{-35.26}{5.29}{0}{neg}{pos}{\opacity}{0} & \midpointgradientcell{-14.93}{-35.26}{5.29}{0}{neg}{pos}{\opacity}{0} & \midpointgradientcell{-14.88}{-35.26}{5.29}{0}{neg}{pos}{\opacity}{0} & \midpointgradientcell{0.42}{-35.26}{5.29}{0}{neg}{pos}{\opacity}{0} & \midpointgradientcell{-2.57}{-35.26}{5.29}{0}{neg}{pos}{\opacity}{0} & \midpointgradientcell{-2.76}{-35.26}{5.29}{0}{neg}{pos}{\opacity}{0} & \midpointgradientcell{-12.42}{-35.26}{5.29}{0}{neg}{pos}{\opacity}{0} & \midpointgradientcell{-0.80}{-35.26}{5.29}{0}{neg}{pos}{\opacity}{0} & \midpointgradientcell{-0.26}{-35.26}{5.29}{0}{neg}{pos}{\opacity}{0} & \midpointgradientcell{-9.15}{-35.26}{5.29}{0}{neg}{pos}{\opacity}{0} & \midpointgradientcell{-6.71}{-35.26}{5.29}{0}{neg}{pos}{\opacity}{0} & \midpointgradientcell{-5.03}{-35.26}{5.29}{0}{neg}{pos}{\opacity}{0} & \midpointgradientcell{-1.67}{-35.26}{5.29}{0}{neg}{pos}{\opacity}{0} \\
\bottomrule
\end{tabular}}
\caption{Effect of perturbations on \textbf{\textit{CheckThat22}} (\(\text{MAP\textit{@}}50\)). \textbf{First-Stage Retrieval} shows the \emph{retrieval gap} between unperturbed and perturbed inputs: $\Delta(\text{MAP}@50)_{\text{retrieval}}$. \textbf{Reranking Recovery} measures the \(\Delta \text{\emph{recovery gap}}\) for the perturbed input set before and after reranking top-\textit{50} candidates, with improvements highlighted in \hlrerank{green} and drops in \hlneg{red}. \textbf{Overall Pipeline} shows the \(\Delta \text{\emph{overall gap}}\) for combined retrieval and reranking between unperturbed and perturbed inputs. Colors indicate \hlpos{positive} or \hlneg{negative} deltas compared to unperturbed. \(\text{MAP\textit{@}}50\) is used for all comparisons. Models finetuned on the task are indicated with the postfix \texttt{ft}.\vspace{-0.5em}}
\label{tab:map@50_checkthat22}
\end{table*}

\subsection{Results for \textit{FactCheckTweet}}
\label{appendix:additional_factcheck_k}
We report evaluation results across different MAP@$k$ values, \( k \in \{5, 10, 20, 50\} \), for the \textit{FactCheckTweet} dataset. Table~\ref{tab:map@5_factcheck} presents the evaluation at MAP@$5$, Table~\ref{tab:map@10_factcheck} at MAP@$10$ and Table~\ref{tab:map@50_factcheck} at MAP@$50$, with reranking applied to the top 50 candidates in all cases.
\begin{table*}[t!]
\centering
\resizebox{0.9\textwidth}{!}{%
\begin{tabular}{ll|cc|cc|cc|cc|cc|cccc}
\toprule
 \multicolumn{2}{l}{} & \multicolumn{2}{c|}{\textbf{Casing}} & \multicolumn{2}{c|}{\textbf{Typos}} & \multicolumn{2}{c|}{\textbf{Negation}} & \multicolumn{2}{c|}{\textbf{Entity Replacement}} & \multicolumn{2}{c|}{\textbf{LLM Rewrite}} & \multicolumn{4}{c}{\textbf{Dialect}} \\
 \textbf{Stage} & \textbf{Model} & \textit{TrueCase} & \textit{Upper} & \textit{Least} & \textit{Most} & \textit{Shallow} & \textit{Double} & \textit{Atleast 1} & \textit{All} & \textit{Least} & \textit{Most} & \textit{AAE} & \textit{Jamaican} & \textit{Pidgin} & \textit{Singlish} \\
\midrule
\parbox[t]{2mm}{\multirow{12}{*}{\rotatebox[origin=c]{90}{
 \parbox[c]{2cm}{\centering  \textbf{First-Stage Retrieval}}}}} 
&\texttt{BM25} & \midpointgradientcell{0.70}{-23.72}{10.71}{0}{neg}{pos}{\opacity}{0} & \midpointgradientcell{0.00}{-23.72}{10.71}{0}{neg}{pos}{\opacity}{0} & \midpointgradientcell{-10.09}{-23.72}{10.71}{0}{neg}{pos}{\opacity}{0} & \midpointgradientcell{-12.40}{-23.72}{10.71}{0}{neg}{pos}{\opacity}{0} & \midpointgradientcell{2.73}{-23.72}{10.71}{0}{neg}{pos}{\opacity}{0} & \midpointgradientcell{6.28}{-23.72}{10.71}{0}{neg}{pos}{\opacity}{0} & \midpointgradientcell{-2.68}{-23.72}{10.71}{0}{neg}{pos}{\opacity}{0} & \midpointgradientcell{-19.95}{-23.72}{10.71}{0}{neg}{pos}{\opacity}{0} & \midpointgradientcell{2.35}{-23.72}{10.71}{0}{neg}{pos}{\opacity}{0} & \midpointgradientcell{-2.33}{-23.72}{10.71}{0}{neg}{pos}{\opacity}{0} & \midpointgradientcell{-6.97}{-23.72}{10.71}{0}{neg}{pos}{\opacity}{0} & \midpointgradientcell{-11.03}{-23.72}{10.71}{0}{neg}{pos}{\opacity}{0} & \midpointgradientcell{-11.50}{-23.72}{10.71}{0}{neg}{pos}{\opacity}{0} & \midpointgradientcell{-2.72}{-23.72}{10.71}{0}{neg}{pos}{\opacity}{0} \\
\bottomrule
\end{tabular}}
\caption{Effect of perturbations on \textbf{\textit{FactCheckTweet}} (\(\text{MAP\textit{@}}5\)). \textbf{First-Stage Retrieval} shows the \emph{retrieval gap} between unperturbed and perturbed inputs: $\Delta(\text{MAP}@5)_{\text{retrieval}}$. \textbf{Reranking Recovery} measures the \(\Delta \text{\emph{recovery gap}}\) for the perturbed input set before and after reranking top-\textit{50} candidates, with improvements highlighted in \hlrerank{green} and drops in \hlneg{red}. \textbf{Overall Pipeline} shows the \(\Delta \text{\emph{overall gap}}\) for combined retrieval and reranking between unperturbed and perturbed inputs. Colors indicate \hlpos{positive} or \hlneg{negative} deltas compared to unperturbed. \(\text{MAP\textit{@}}5\) is used for all comparisons. Models finetuned on the task are indicated with the postfix \texttt{ft}.\vspace{-0.5em}}
\label{tab:map@5_factcheck}
\end{table*}

\begin{table*}[t!]
\centering
\resizebox{0.9\textwidth}{!}{%
\begin{tabular}{ll|cc|cc|cc|cc|cc|cccc}
\toprule
 \multicolumn{2}{l}{} & \multicolumn{2}{c|}{\textbf{Casing}} & \multicolumn{2}{c|}{\textbf{Typos}} & \multicolumn{2}{c|}{\textbf{Negation}} & \multicolumn{2}{c|}{\textbf{Entity Replacement}} & \multicolumn{2}{c|}{\textbf{LLM Rewrite}} & \multicolumn{4}{c}{\textbf{Dialect}} \\
 \textbf{Stage} & \textbf{Model} & \textit{TrueCase} & \textit{Upper} & \textit{Least} & \textit{Most} & \textit{Shallow} & \textit{Double} & \textit{Atleast 1} & \textit{All} & \textit{Least} & \textit{Most} & \textit{AAE} & \textit{Jamaican} & \textit{Pidgin} & \textit{Singlish} \\
\midrule
\parbox[t]{2mm}{\multirow{12}{*}{\rotatebox[origin=c]{90}{
 \parbox[c]{2cm}{\centering  \textbf{First-Stage Retrieval}}}}} 
&\texttt{BM25} & \midpointgradientcell{0.68}{-23.78}{10.70}{0}{neg}{pos}{\opacity}{0} & \midpointgradientcell{0.00}{-23.78}{10.70}{0}{neg}{pos}{\opacity}{0} & \midpointgradientcell{-9.31}{-23.78}{10.70}{0}{neg}{pos}{\opacity}{0} & \midpointgradientcell{-12.33}{-23.78}{10.70}{0}{neg}{pos}{\opacity}{0} & \midpointgradientcell{2.51}{-23.78}{10.70}{0}{neg}{pos}{\opacity}{0} & \midpointgradientcell{5.93}{-23.78}{10.70}{0}{neg}{pos}{\opacity}{0} & \midpointgradientcell{-2.92}{-23.78}{10.70}{0}{neg}{pos}{\opacity}{0} & \midpointgradientcell{-19.50}{-23.78}{10.70}{0}{neg}{pos}{\opacity}{0} & \midpointgradientcell{2.65}{-23.78}{10.70}{0}{neg}{pos}{\opacity}{0} & \midpointgradientcell{-2.40}{-23.78}{10.70}{0}{neg}{pos}{\opacity}{0} & \midpointgradientcell{-6.71}{-23.78}{10.70}{0}{neg}{pos}{\opacity}{0} & \midpointgradientcell{-10.42}{-23.78}{10.70}{0}{neg}{pos}{\opacity}{0} & \midpointgradientcell{-11.27}{-23.78}{10.70}{0}{neg}{pos}{\opacity}{0} & \midpointgradientcell{-2.54}{-23.78}{10.70}{0}{neg}{pos}{\opacity}{0} \\
\bottomrule
\end{tabular}}
\caption{Effect of perturbations on \textbf{\textit{FactCheckTweet}} (\(\text{MAP\textit{@}}10\)). \textbf{First-Stage Retrieval} shows the \emph{retrieval gap} between unperturbed and perturbed inputs: $\Delta(\text{MAP}@10)_{\text{retrieval}}$. \textbf{Reranking Recovery} measures the \(\Delta \text{\emph{recovery gap}}\) for the perturbed input set before and after reranking top-\textit{50} candidates, with improvements highlighted in \hlrerank{green} and drops in \hlneg{red}. \textbf{Overall Pipeline} shows the \(\Delta \text{\emph{overall gap}}\) for combined retrieval and reranking between unperturbed and perturbed inputs. Colors indicate \hlpos{positive} or \hlneg{negative} deltas compared to unperturbed. \(\text{MAP\textit{@}}10\) is used for all comparisons. Models finetuned on the task are indicated with the postfix \texttt{ft}.\vspace{-0.5em}}
\label{tab:map@10_factcheck}
\end{table*}

\begin{table*}[t!]
\centering
\resizebox{0.9\textwidth}{!}{%
\begin{tabular}{ll|cc|cc|cc|cc|cc|cccc}
\toprule
 \multicolumn{2}{l}{} & \multicolumn{2}{c|}{\textbf{Casing}} & \multicolumn{2}{c|}{\textbf{Typos}} & \multicolumn{2}{c|}{\textbf{Negation}} & \multicolumn{2}{c|}{\textbf{Entity Replacement}} & \multicolumn{2}{c|}{\textbf{LLM Rewrite}} & \multicolumn{4}{c}{\textbf{Dialect}} \\
 \textbf{Stage} & \textbf{Model} & \textit{TrueCase} & \textit{Upper} & \textit{Least} & \textit{Most} & \textit{Shallow} & \textit{Double} & \textit{Atleast 1} & \textit{All} & \textit{Least} & \textit{Most} & \textit{AAE} & \textit{Jamaican} & \textit{Pidgin} & \textit{Singlish} \\
\midrule
\parbox[t]{2mm}{\multirow{12}{*}{\rotatebox[origin=c]{90}{
 \parbox[c]{2cm}{\centering  \textbf{First-Stage Retrieval}}}}}
&\texttt{BM25} & \midpointgradientcell{0.66}{-23.82}{10.45}{0}{neg}{pos}{\opacity}{0} & \midpointgradientcell{0.00}{-23.82}{10.45}{0}{neg}{pos}{\opacity}{0} & \midpointgradientcell{-9.09}{-23.82}{10.45}{0}{neg}{pos}{\opacity}{0} & \midpointgradientcell{-11.87}{-23.82}{10.45}{0}{neg}{pos}{\opacity}{0} & \midpointgradientcell{2.51}{-23.82}{10.45}{0}{neg}{pos}{\opacity}{0} & \midpointgradientcell{5.94}{-23.82}{10.45}{0}{neg}{pos}{\opacity}{0} & \midpointgradientcell{-2.52}{-23.82}{10.45}{0}{neg}{pos}{\opacity}{0} & \midpointgradientcell{-19.21}{-23.82}{10.45}{0}{neg}{pos}{\opacity}{0} & \midpointgradientcell{2.63}{-23.82}{10.45}{0}{neg}{pos}{\opacity}{0} & \midpointgradientcell{-2.12}{-23.82}{10.45}{0}{neg}{pos}{\opacity}{0} & \midpointgradientcell{-6.71}{-23.82}{10.45}{0}{neg}{pos}{\opacity}{0} & \midpointgradientcell{-10.22}{-23.82}{10.45}{0}{neg}{pos}{\opacity}{0} & \midpointgradientcell{-11.09}{-23.82}{10.45}{0}{neg}{pos}{\opacity}{0} & \midpointgradientcell{-2.39}{-23.82}{10.45}{0}{neg}{pos}{\opacity}{0} \\
\bottomrule
\end{tabular}}
\caption{Effect of perturbations on \textbf{\textit{FactCheckTweet}} (\(\text{MAP\textit{@}}50\)). \textbf{First-Stage Retrieval} shows the \emph{retrieval gap} between unperturbed and perturbed inputs: $\Delta(\text{MAP}@50)_{\text{retrieval}}$. \textbf{Reranking Recovery} measures the \(\Delta \text{\emph{recovery gap}}\) for the perturbed input set before and after reranking top-\textit{50} candidates, with improvements highlighted in \hlrerank{green} and drops in \hlneg{red}. \textbf{Overall Pipeline} shows the \(\Delta \text{\emph{overall gap}}\) for combined retrieval and reranking between unperturbed and perturbed inputs. Colors indicate \hlpos{positive} or \hlneg{negative} deltas compared to unperturbed. \(\text{MAP\textit{@}}50\) is used for all comparisons. Models finetuned on the task are indicated with the postfix \texttt{ft}.\vspace{-0.5em}}
\label{tab:map@50_factcheck}
\end{table*}

\section{Mitigation Approaches Comparison}
\begin{figure*}[h!]
    \centering
    \includegraphics[width=1.0\textwidth]{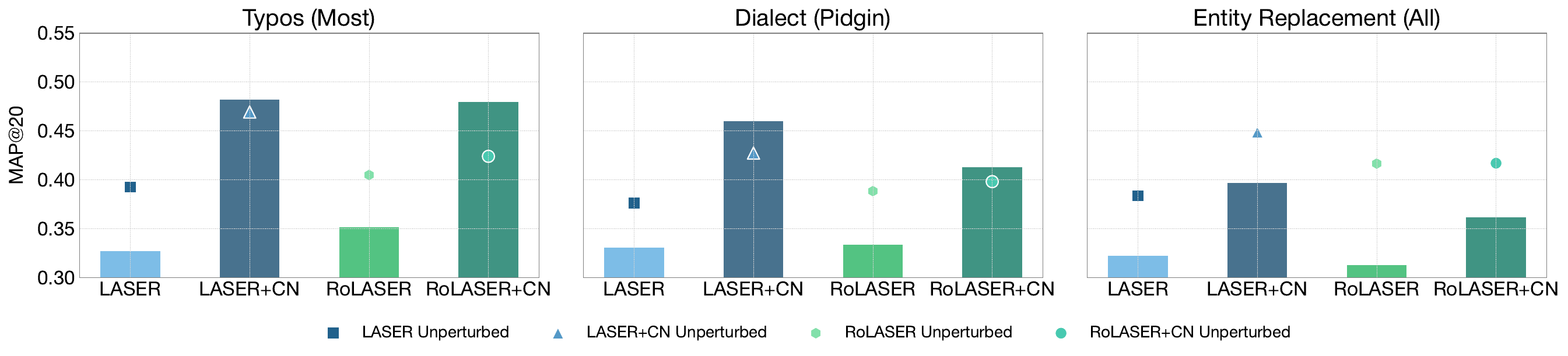}
    \vspace{-4mm}
    \caption{Effect of mitigation approaches on worst-case perturbations for \textit{typos}, \textit{dialect}, and \textit{entity replacement}. For each embedding model, markers represent performance on the \hlbase{unperturbed} input set, while the bars indicate performance on the \hlpet{perturbed} set. \texttt{RoLASER} is a robust-\texttt{LASER} model adapted for user-generated content~\cite{nishimwe-etal-2024-making}, while models denoted with +CN incorporate \textit{claim normalization}.
}
    \label{fig:mitigation_results}
\end{figure*}

We compare our mitigation approaches on the Robust-LASER (\texttt{RoLASER}) model proposed by \citet{nishimwe-etal-2024-making}, who use a similar teacher–student knowledge distillation approach to improve the robustness of embedding models for user-generated content (UGC). In contrast to our work, they focus only on a subset of perturbations—specifically, those in which UGC resembles the \textit{Typos} perturbation type—and generate an artificial UGC dataset using rule-based methods. For the KD approach, we compare the performance of the original \texttt{LASER} model with that of Robust-LASER (\texttt{RoLASER}), the adapted model. We also evaluate the effect of \textit{claim normalization} using the same setup as in our approach. The results for this experiment are shown in Figure~\ref{fig:mitigation_results}. 

Although the mitigation approaches are not directly comparable here due to the use of different embedding models (i.e., \texttt{LASER} vs.\ \texttt{all-mpnet-base-v2}), we observe that \textit{claim normalization} is generally beneficial across all perturbation types. Additionally, \texttt{RoLASER} outperforms the standard \texttt{LASER} model on typos and shows marginal improvement on dialect perturbations. We hypothesize that this occurs because \texttt{RoLASER} is trained exclusively on user-generated content and not on other variations (e.g., \textit{dialect} or \textit{entity replacements}), which limits its generalizability to perturbation types not included in the augmentation set. Finally, we note that the LASER models generally perform poorly on this task compared to the other embedding models we evaluate.

\section{Scaling Laws and Performance Drop}
\begin{figure*}
    \centering
    \includegraphics[width=1\textwidth]{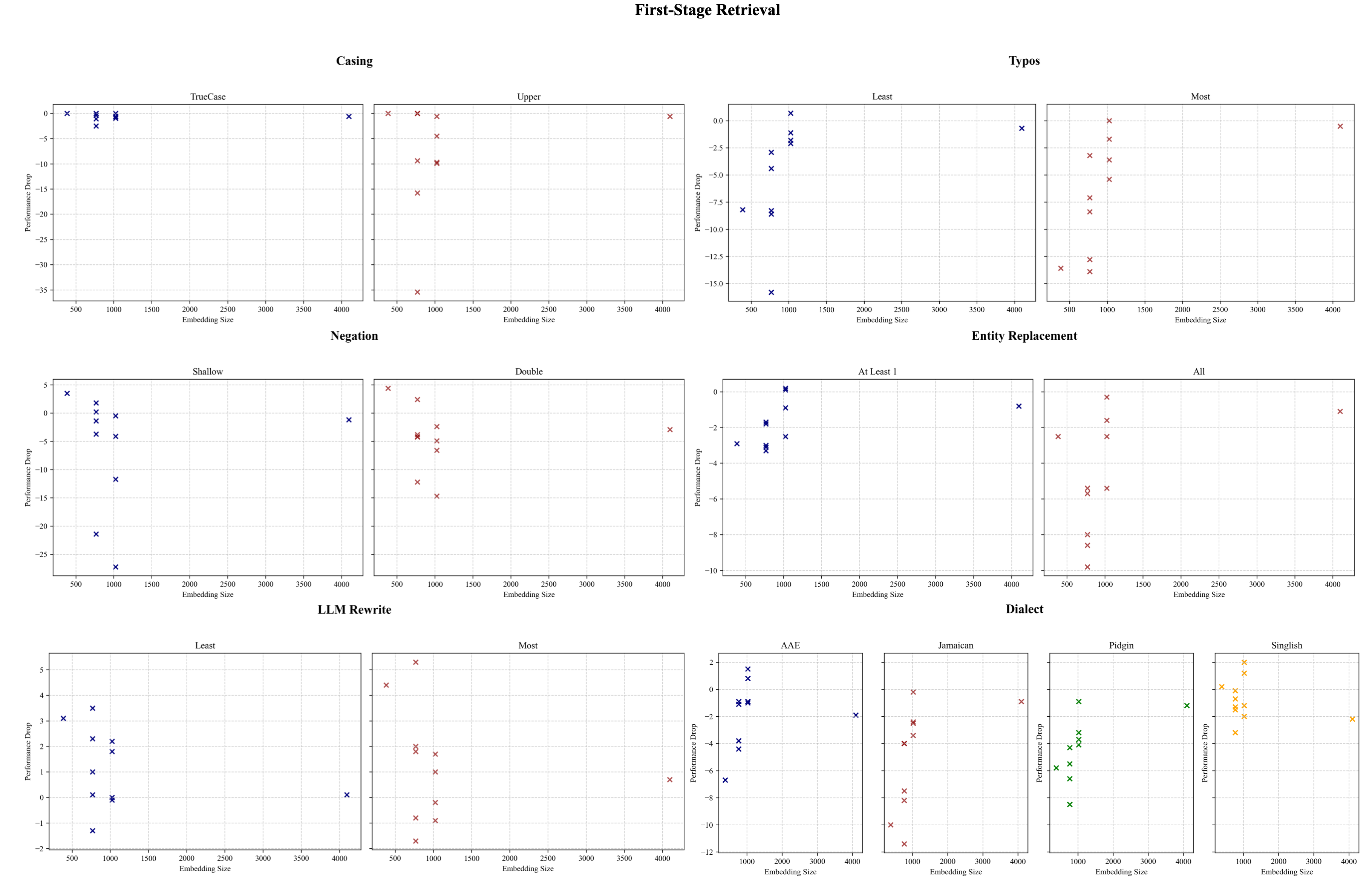}
    \caption{Correlation between a model embedding's size and the performance gap for the First-Stage Retrieval as reported in Table~\ref{tab:map@20_checkthat}.}\label{fig:scatter-first-stage}
\end{figure*}

\begin{figure*}
    \centering
    \includegraphics[width=1\textwidth]{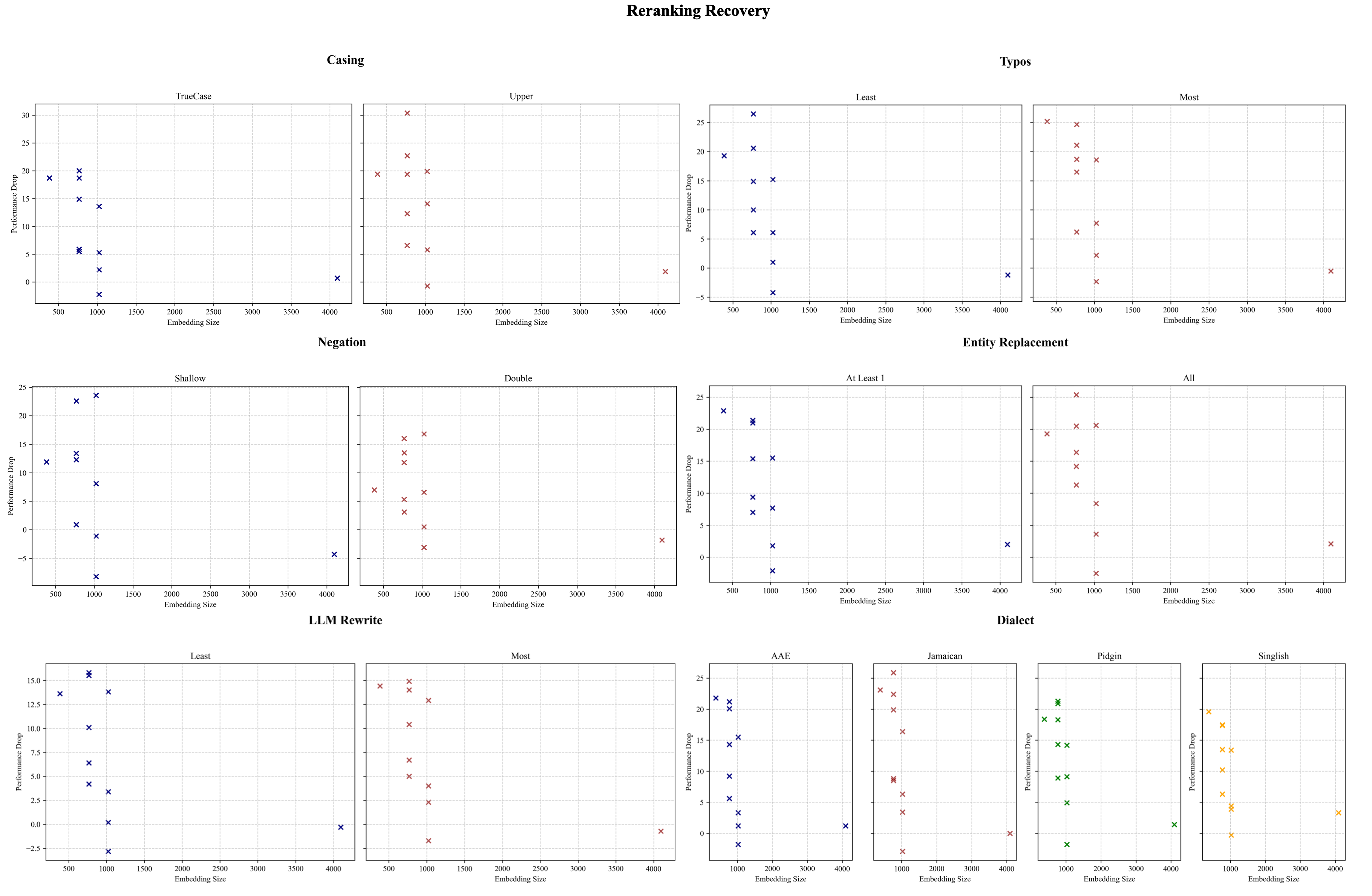}
    \caption{Correlation between a model embedding's size and the performance gap for the Reranking Recovery as reported in Table~\ref{tab:map@20_checkthat}.}\label{fig:scatter-reranking}
\end{figure*}

\begin{figure*}
    \centering
    \includegraphics[width=1\textwidth]{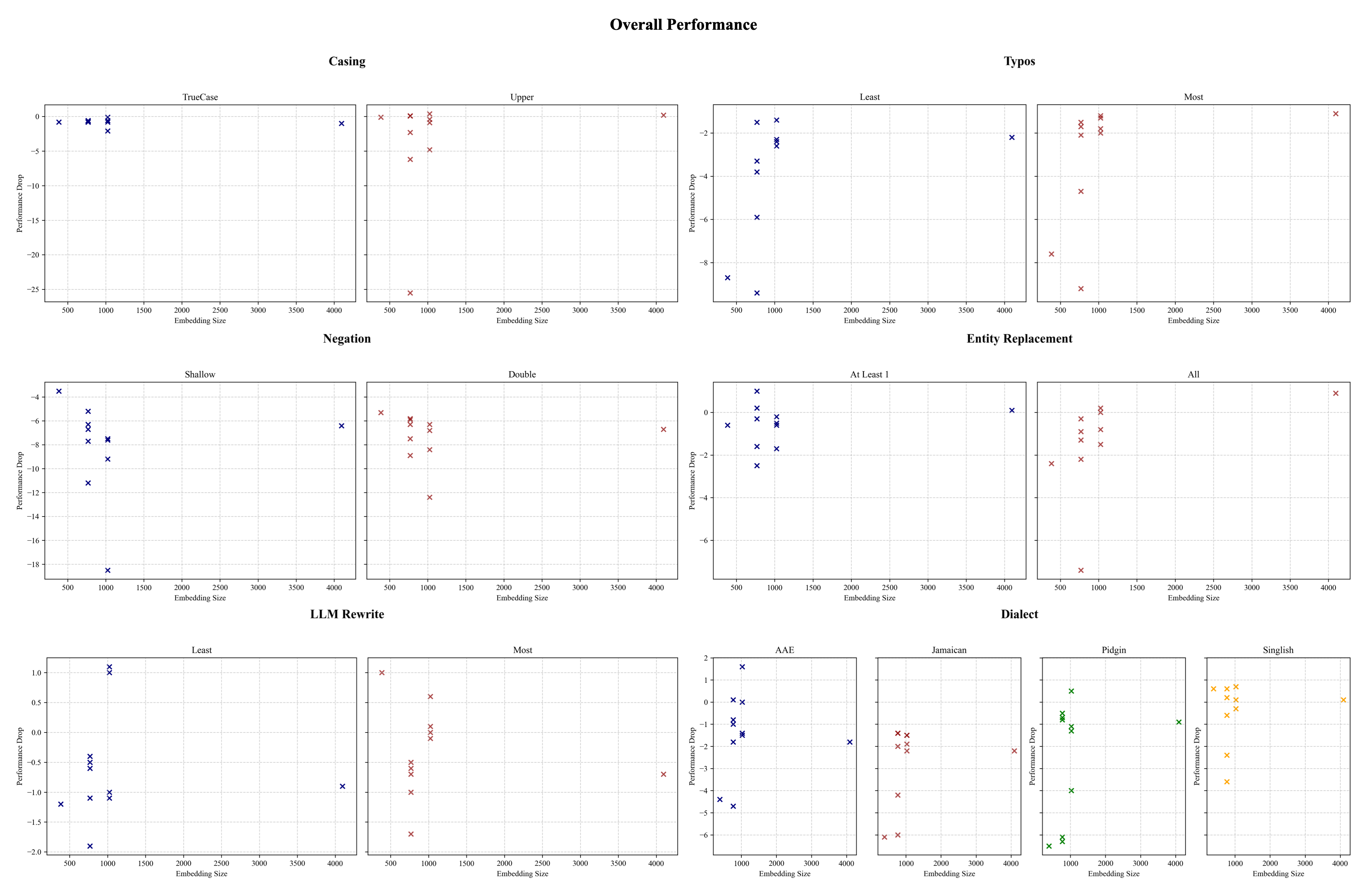}
    \caption{Correlation between a model embedding's size and the performance gap for the Overall Performance as reported in Table~\ref{tab:map@20_checkthat}.}\label{fig:scatter-overall}
\end{figure*}

\begin{figure*}
    \centering
    \includegraphics[width=1\textwidth]{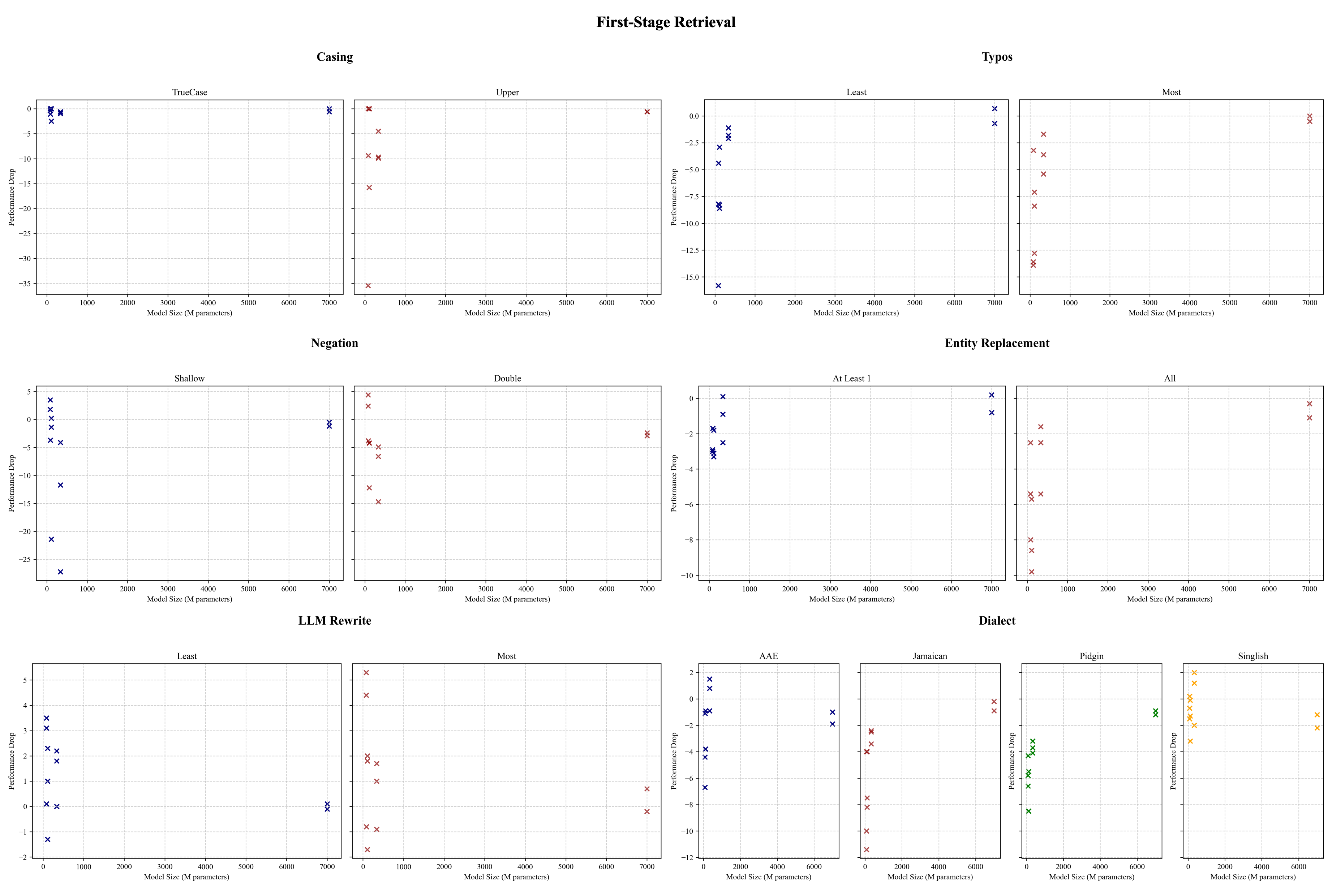}
    \caption{Correlation between model size (millions of parameters) and the performance gap for the First-Stage Retrieval as reported in Table~\ref{tab:map@20_checkthat}.}\label{fig:scatter-first-stage-size}
\end{figure*}

\begin{figure*}
    \centering
    \includegraphics[width=1\textwidth]{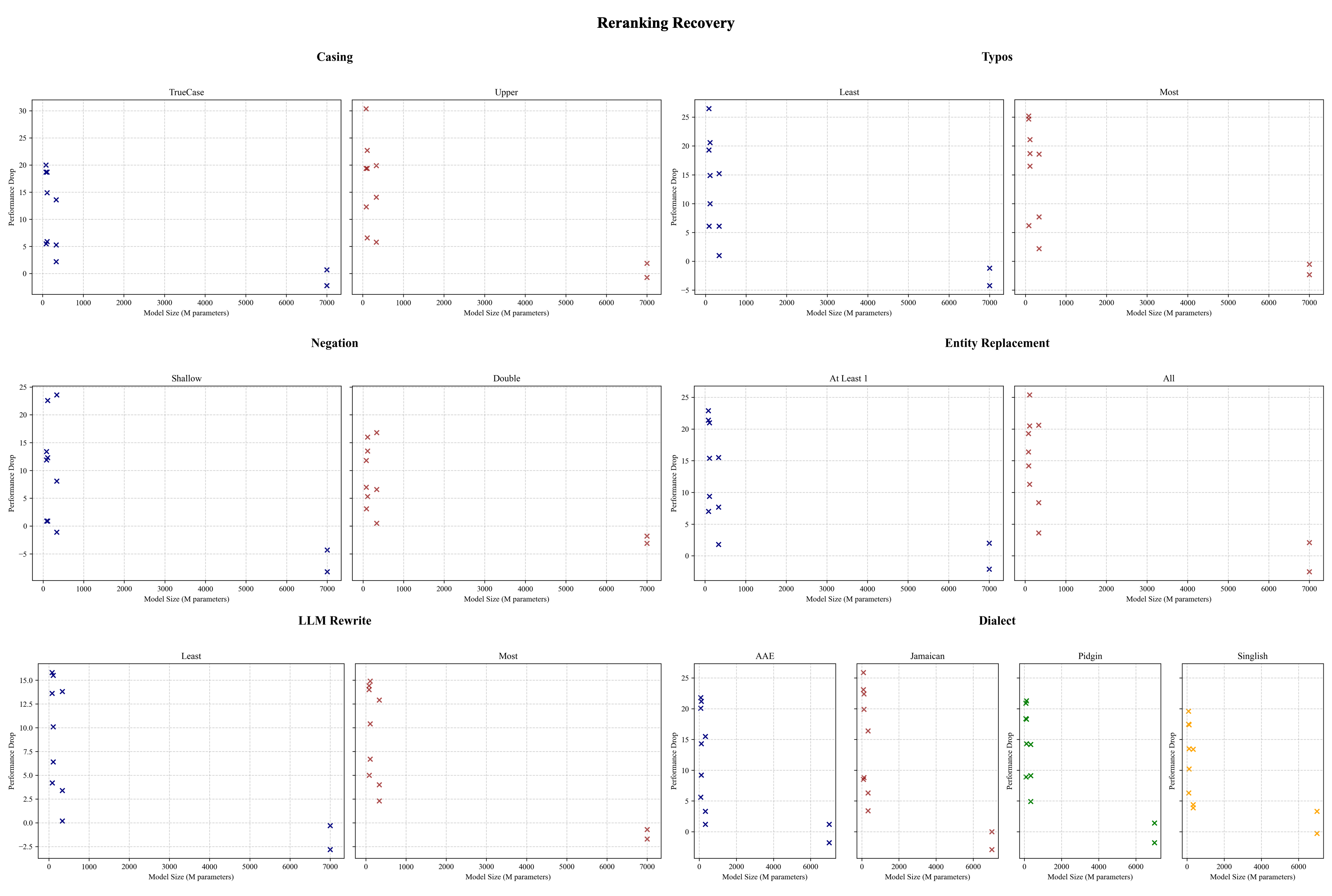}
    \caption{Correlation between model size (millions of parameters) and the performance gap for the Reranking Recovery as reported in Table~\ref{tab:map@20_checkthat}.}\label{fig:scatter-reranking-size}
\end{figure*}

\begin{figure*}
    \centering
    \includegraphics[width=1\textwidth]{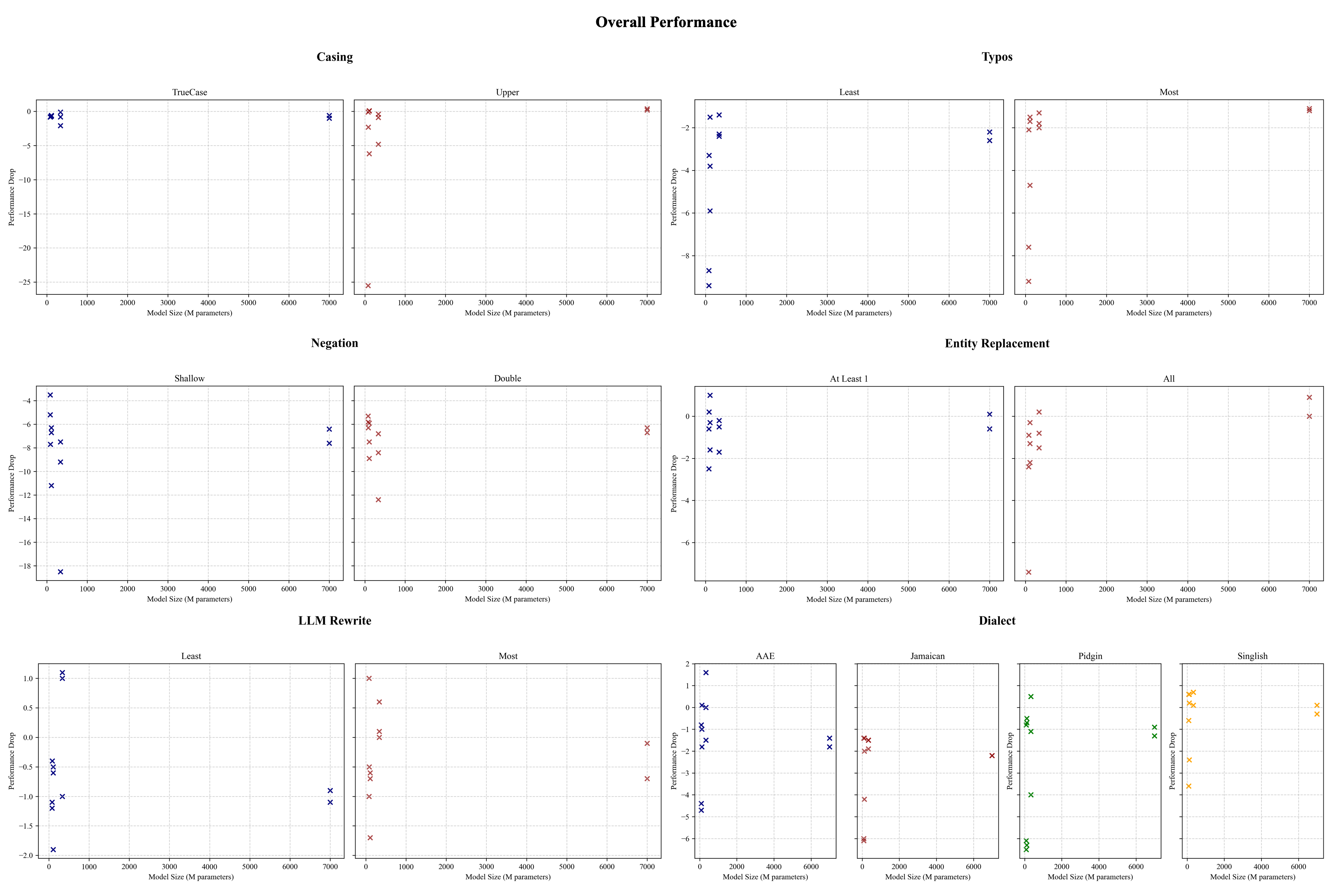}
    \caption{Correlation between model size (millions of parameters) and the performance gap for the Overall Performance as reported in Table~\ref{tab:map@20_checkthat}.}\label{fig:scatter-overall-size}
\end{figure*}

We report an analysis of the performance gap as reported in Table~\ref{tab:map@20_checkthat}, and the models' size and embedding used to represent the sentences at each stage.
Figures~\ref{fig:scatter-first-stage}, ~\ref{fig:scatter-first-stage-size} reports the correlation for the First-Stage Retrieval (embedding and model size); Figures~\ref{fig:scatter-reranking} and~\ref{fig:scatter-reranking-size} that for Reranking Recovery (embedding and model size); Figures~\ref{fig:scatter-overall} and~\ref{fig:scatter-overall-size} the Overall Pipeline (embedding and model size). We excluded \texttt{BM25} from the analysis as there is no proper notion of embedding dimension for that technique. In general, we do not observe any significant correlation between the embedding dimension and the performance gap.

\onecolumn
\clearpage
\section{Prompts}
\label{appendix:prompt}
We provide examples of prompts used in the perturbation generation framework, including the \textit{LLM as a perturber} and \textit{LLM as a verifier} prompts, as well as the \textit{claim normalization} prompt used in the mitigation approaches. For brevity, we only present prompts for \textit{Dialect} perturbation and verification; all other prompts are available in the repository. To encourage diversity in the generated perturbations, we set the generation temperature to \(0.9\), while for verification and claim normalization, we use a temperature of \(0\) to ensure reproducibility. 

\begin{prompt}[title={Dialect Perturbation}]
You are now a social media user tasked with rewriting a given tweet in different English dialects. You will receive two inputs:
\begin{enumerate}
    \item A claim (original tweet)
    \item A fact-check that supports or refutes the claim
\end{enumerate}

\textbf{Your task:} Rewrite the given tweet in as many different ways as possible using the following dialects: African American Vernacular English, Nigerian Pidgin English, Singlish (Singapore English), Jamaican Patois

\textbf{Ensure that:}
\begin{itemize}
    \item The fact-check remains applicable to the rewritten tweets.
    \item The rewritten tweets convey the same main claim as the original.
    \item The rewritten tweets read naturally and do not appear suspiciously altered.
\end{itemize}

\textbf{Example:}\\[1mm]
\textbf{Original tweet:} ``Biden signed an executive order today banning the term 'China virus'."\\[1mm]
\textbf{Fact-check:} ``President Joe Biden issued an executive order in January 2021 banning the term ‘China virus.’ ''\\[1mm]
\textbf{Possible rewrites:}
\begin{itemize}
    \item Biden done signed an order today, no more sayin’ ‘China virus.’
    \item Biden don sign executive order today, e no go dey call am ‘China virus’ again.
    \item Wah, Biden just sign order today, cannot call ‘China virus’ liao.
    \item Biden sign one order today, fi stop di use a ‘China virus.’
\end{itemize}

\textbf{Response Format:}
\begin{itemize}
    \item Rewritten Tweet 1: [Your first rewritten version]
    \item Rewritten Tweet 2: [Your second rewritten version]
    \item \ldots (and so on)
\end{itemize}

\textbf{Inputs:}\\[1mm]
Tweet: \{claim\}\\[1mm]
Fact Check: \{fact\_check\}

Generate your response in the specified string format.
\end{prompt}

\begin{prompt}[title={Dialect Verification}]
You are now a fact checker tasked with verifying whether a fact-check is applicable to a list of rewritten tweets. You will receive three inputs:
\begin{enumerate}
    \item Fact-check: A statement supporting or refuting a claim.
    \item Original Tweet: The source tweet conveying the claim.
    \item Rewritten Tweets: A list of tweets rewritten based on the original tweet in different English dialects.
\end{enumerate}

\textbf{Your task:} For each rewritten tweet, evaluate:
\begin{itemize}
    \item Does the fact-check apply to it (i.e., is the fact-check helpful in verifying the rewritten tweet)?
    \item Does it convey the same main claim as the original tweet?
    \item Does the rewritten tweet read naturally, as if written by a typical social media user?
\end{itemize}

\textbf{Your output:} For each rewritten tweet, provide a binary label indicating whether the constraints above are satisfied.

\textbf{Example:}\\[2mm]
\textbf{Fact-check:} ``President Joe Biden issued an executive order in January 2021 banning the term ‘China virus.’''\\[1mm]
\textbf{Original tweet:} ``Biden signed an executive order today banning the term `China virus'.''\\[2mm]
\textbf{Rewritten tweets:}
\begin{itemize}[nosep,leftmargin=*]
    \item Biden done signed an order today, no more sayin’ ‘China virus.’
    \item Biden don sign executive order today, e no go dey call am ‘China virus’ again.
    \item Wah, Biden just sign order today, cannot call ‘China virus’ liao.
    \item Biden sign one order today, fi stop di use a `Bejing virus.’
\end{itemize}
\textbf{Output:}
\begin{verbatim}
{
    "labels": [1, 1, 1, 0]
}
\end{verbatim}

\textbf{Inputs:}\\[1mm]
\textbf{Fact Check:} \{fact\_check\}\\[1mm]
\textbf{Original Tweet:} \{claim\}\\[1mm]
\textbf{Rewritten Tweets:} \{rewrites\}\\[1mm]

Generate your response in the specified JSON format.
\end{prompt}

\begin{prompt}[title={Claim Normalization}]
You will be provided with a noisy input claim from a social media post. The input claim may contain informal language, typos, abbreviations, double negations, and dialectal variations. Your task is to normalise the claim to a more formal and standardised version while preserving the original meaning.

\textbf{Ensure that:}
\begin{itemize}
    \item The normalised claim conveys the same main claim as the original.
\end{itemize}

\textbf{Let's see an example:}

\textbf{Noisy Claim:} ``Wah, Biden just sign order today, cannot call ‘China virus’ liao''\\[1mm]
\textbf{Normalised Claim:} ``President Joe Biden issued an executive order today banning the term ‘China virus.’''

\textbf{Noisy Claim:} ``Soros son sez he and dad pickd Harris 4 VP after pic interview!''\\[1mm]
\textbf{Normalised Claim:} ``George Soros son revealed that he and his father chose Kamala Harris as the Vice President after a picture interview.''

\textbf{Noisy Claim:} ``It is not untrue that President-elect Joe Biden’s German shepherd, Major, is set to become the first shelter dog in the White House.''\\[1mm]
\textbf{Normalised Claim:} ``President-elect Joe Biden’s German shepherd, Major, is set to become the first shelter dog in the White House.''

\textbf{Response Format:}\\
Normalised Claim: [Your normalised claim]

\textbf{Inputs:}\\
Noisy Claim: \{claim\}

Generate your response in the specified string format.
\end{prompt}

\end{document}